\documentclass[10pt,twocolumn,letterpaper]{article}

\usepackage[pagenumbers]{wacv} 
\usepackage{multirow}
\usepackage{tabularx}
\usepackage{array}
\usepackage[dvipsnames,table,xcdraw]{xcolor}
\definecolor{wacvblue}{rgb}{0.21,0.49,0.74}
\usepackage[pagebackref,breaklinks,colorlinks,allcolors=wacvblue]{hyperref}

\def\wacvPaperID{1101} 
\def\confName{WACV}
\def\confYear{2027}

\title{MATA: Motion-aligned Temporal Adapter}

\author{
\begin{tabular}[t]{c}
Ashish Kumar Singh$^{*}$\\
Machine Learning Group\\
Luleå University of Technology, Sweden\\
{\tt\small ashish.kumar.singh@associated.ltu.se}
\end{tabular}
\hspace{3em}
\begin{tabular}[t]{c}
Prakash Chandra Chhipa$^{*}$\\
Scalable Systems\\
RISE Research Institutes of Sweden, Sweden\\
{\tt\small prakash.chandra.chhipa@ri.se}
\end{tabular}
\\[8pt]
\textit{$^{*}$Joint first authors with equal contributions.}
}

\begin{document}

\maketitle

\begin{abstract}

Few-shot fine-grained video action recognition requires recognizing novel action classes from only a few labeled examples. The difficulty is that visually similar actions often share actors, scenes, poses, and objects, differing only in the order, timing, and short duration of motion changes. We study this problem under three practical constraints: few tunable parameters, low compute cost, and use of the video modality alone. We propose \textbf{Motion-aligned Temporal Adapter (MATA)}, a parameter-efficient adapter on top of a frozen CLIP backbone. MATA specializes CLIP features along the temporal dimension using two components: \textit{temporal phase alignment} through learnable temporal positional embeddings, and a \textit{local motion transition operator} through depth-wise convolution. With only $\sim$3.6M tunable parameters, MATA achieves state-of-the-art performance across multiple fine-grained benchmarks while remaining robust to both visual corruptions and temporal disruptions induced by frame dropping. Code is available at \url{https://github.com/prakashchhipa/MATA-Motion-aligned-Temporal-Adapter}.


\end{abstract}    
\begin{figure*}[h]
  \centering
  \includegraphics[width=1.0\linewidth]{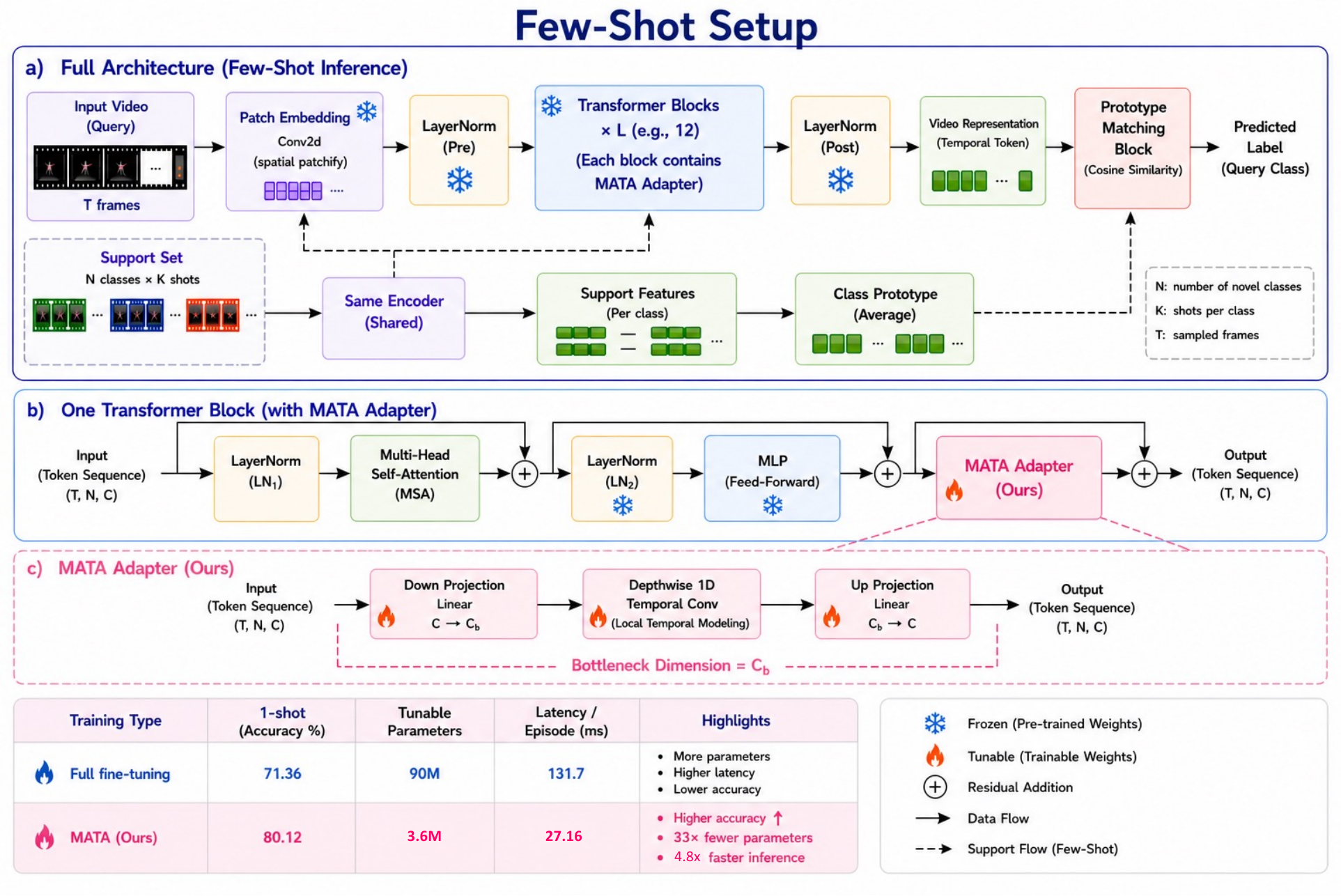}
\caption{
\textbf{Overview of the MATA few-shot inference pipeline}. Support and query videos are encoded by a shared frozen CLIP backbone, and query labels are predicted via prototype matching. Compared to full fine-tuning \cite{wang2024clip}, MATA updates only a small set of parameters, improving accuracy while reducing latency.
}
  \label{fig:overview_and_results1}
\end{figure*}

\section{Introduction}

Few-shot fine-grained action recognition requires separating novel actions given only a few labeled examples.
The challenge is that visually similar actions often share actors, scenes, poses, and objects \cite{li2018diving48, shao2020finegym}, while differing in brief motion changes, timing, and action phases (e.g., preparation, execution, and completion stages). Thus, recognition is both low-data and temporally sensitive: support-query matching must align action phases and specialize temporal features to capture the motion cues that separate fine-grained classes.

Existing few-shot action recognition methods face two practical limitations. Metric- and alignment-based methods such as OTAM~\citep{cao2020few}, TRX~\citep{perrett2021temporal}, STRM~\citep{thatipelli2022spatio}, and HyRSM~\citep{wang2022hybrid} improve support-query matching, but lack the transferable priors of foundation models. CLIP-based methods address this limitation, yet often rely on full backbone fine-tuning, additional modalities as in CLIP-FSAR \cite{wang2024clip}, MA-FSAR \cite{xing2026ma}, or heavier meta-learning designs~\citep{wang2024clip,xing2026ma}. On the other hand, parameter-efficient video adapters on CLIP reduce tunable parameters, but typically combine spatial and temporal adaptation through multi-branch designs (AIM \cite{yang2023aim}), attention-based modules (D$^2$ST \cite{pei2025d2st}), or 3D convolutions (ST-Adapter ~\citep{pan2022st}). This can add computation and can dilute the motion-specific cues that are most critical in few-shot fine-grained action recognition. A third line of work improves temporal reasoning through explicitly extracted motion structure: Trokens~\cite{kumar2025trokens} builds trajectory tokens in a separate offline stage before training, which adds preprocessing time and storage, and makes recognition sensitive to the quality of the extracted tokens.

This leads to the research question: \textit{Can CLIP be adapted for few-shot fine-grained video action recognition using a minimal, compute-efficient mechanism that focuses on temporal dynamics while using only the video modality?} Therefore, we study few-shot fine-grained video action recognition under the accuracy--compute efficiency--unimodality objective: i) improving recognition performance with few tunable parameters, ii) being compute efficient, and iii) using only the readily available video modality. This objective follows directly from the challenges above. The model must preserve CLIP's transferable spatial and semantic priors, recover discriminative temporal evidence from support and query videos, and avoid the cost of full backbone tuning, heavy spatio-temporal adaptation, or additional modality pipelines.

We propose \textbf{M}otion-\textbf{a}ligned \textbf{T}emporal \textbf{A}dapter (\textbf{MATA} in Fig \ref{fig:overview_and_results1}), a parameter-efficient adapter for a frozen CLIP visual backbone. MATA is motivated by two needs in few-shot fine-grained recognition: support-query action phases must be aligned, and brief motion transitions must be modeled explicitly. MATA addresses the first need through \textit{temporal phase alignment} (i.e., representing each frame with its relative position in the action sequence), achieved by adding learnable temporal positional embeddings to patch-level bottleneck features so that each visual state is tied to its relative action phase. It addresses the second need through \textit{local motion transition modeling}, achieved by depth-wise one-dimensional temporal convolution along the frame dimension to capture fine-grained details. The resulting temporal response is projected back to the CLIP token dimension through identity-preserving residual reconstruction, keeping adaptation compact while preserving CLIP's pretrained priors.

With only $\sim$3.6M tunable parameters, MATA sets a new state of the art on fine-grained few-shot benchmarks under a CLIP backbone, reaching $80.12\%$ and $81.28\%$ 1-shot accuracy on Diving48 and FineGym288 and improving over D$^2$ST~\cite{pei2025d2st} by $6.83$ and $+10.95$ points while reducing per-episode latency by $4.10\times$. MATA also remains the most accurate method under all five test-time corruptions applied to query clips, showing that focused temporal adaptation can be accurate, compact, fast, and robust from the video modality alone. Our contributions are:
\begin{itemize}
     \item We show that temporal-only adaptation is sufficient for few-shot fine-grained action recognition, effectively capturing short-lived and rapid motion differences between visually similar actions without adapting spatial features.
   
    \item We introduce MATA, a parameter-efficient temporal adapter that adapts a frozen CLIP visual backbone for few-shot recognition from short video clips, requiring neither additional modalities nor auxiliary cues such as pose or offline preprocessing, while substantially reducing tunable parameters and computational cost.
   
    \item We establish MATA’s effectiveness and robustness through extensive evaluation across fine-grained action benchmarks, achieving state-of-the-art few-shot recognition while remaining robust to both visual corruptions and temporal disruptions induced by frame dropping.
\end{itemize}

\section{Related Work}
\label{sec:relwork}

\noindent\textbf{Vision--language pretraining for video actions.}
CLIP~\citep{radford2021clip} provides open-vocabulary visual representations that have been extended to video by ActionCLIP~\citep{xu2021actionclip}, X-CLIP~\citep{ni2022expanding}, ViFi-CLIP~\citep{rasheed2023fine}, and CAST~\citep{lee2023cast} through video--text alignment, cross-frame attention, temporal alignment, and spatial--temporal cross-attention. These rely on full backbone fine-tuning or heavy video-specific modules, which is inefficient for fine-grained actions with rapid, subtle motion.

\noindent\textbf{Parameter-efficient video adaptation.}
Following lightweight adapters in NLP~\citep{houlsby2019parameter}, ST-Adapter~\citep{pan2022st}, DUALPATH~\citep{park2023dual}, D$^2$ST~\citep{pei2025d2st}, and AIM~\citep{yang2023aim} adapt ViT backbones through spatial--temporal modules built from multi-branch or attention-based designs, while Task-Adapter~\citep{cao2024task} and Task-Adapter++~\citep{cao2026task} add task-specific and order-aware alignment. All jointly model spatial appearance and temporal structure. We instead target fine-grained recognition, where short-duration motion is the main discriminative signal, through focused temporal adaptation without additional spatial modeling.

\noindent\textbf{Few-shot action recognition.}
Classical FSAR methods such as ProtoNet~\citep{snell2017prototypical}, OTAM~\citep{cao2020few}, TRX~\citep{perrett2021temporal}, PAL~\citep{zhu2021few}, STRM~\citep{thatipelli2022spatio}, and HyRSM~\citep{wang2022hybrid} learn global similarity metrics or temporal alignments, while VPD~\citep{hong2021video} distills pose supervision into RGB features. Such cues are less suited to actions separated by brief motion evolution rather than coarse appearance. CLIP-based methods exploit stronger semantic priors: CLIP-FSAR~\citep{wang2024clip} fine-tunes the full backbone with temporal transformers, and MA-FSAR~\citep{xing2026ma} freezes it but adds multi-modal and meta-learning components. Others refine matching or temporal structure: TEAM~\citep{lee2025temporal} removes explicit temporal alignment, MPL~\citep{yang2026mpl} learns match-guided prototypes, TSAM~\citep{li2025frame} models frame order as a sequence, HMP-Net~\citep{hmpnet} builds hierarchical meta-prototypes, and EMP-Net~\citep{wu2024efficient} reduces cost via multi-level post reasoning. TATs~\citep{kumar2024trajectory} and Trokens~\citep{kumar2025trokens} instead represent motion as trajectory tokens, which is effective on fine-grained data but requires an offline extraction stage and depends on tracker quality.

MATA retains a frozen CLIP visual backbone and introduces a minimal temporal-only adapter, avoiding attention-heavy or multi-branch architectures, auxiliary modalities, and offline preprocessing, while achieving strong performance on FineGym99, FineGym288, and Diving48.
\section{Method}
\label{sec:method}

\subsection{Problem setting}
\label{sec:method_problem}

We consider $N$-way $K$-shot fine-grained action recognition under the episodic setting. 
Each episode contains a support set
\begin{equation}
\mathcal{S}
=
\{(v_i,y_i)\}_{i=1}^{NK},
\qquad
y_i\in\{1,\ldots,N\},
\label{eq:support_set}
\end{equation}
and a query set
\begin{equation}
\mathcal{Q}
=
\{v_j^{q}\}_{j=1}^{M}.
\label{eq:query_set}
\end{equation}

The support set provides $K$ labeled videos for each of the $N$ action classes. 
For each query video $v_j^{q}$, the recognition objective is to predict
\begin{equation}
\hat{y}_j^{q}\in\{1,\ldots,N\}
\label{eq:query_prediction}
\end{equation}
using only the labeled support videos from the same episode. Videos are encoded by a shared frozen CLIP visual backbone with $T$ segment-sampled
frames, so index $t$ denotes a normalized position within the action. A video is
represented by the mean of its frame-level class tokens, $z(v)=\frac{1}{T}\sum_t
x_{\mathrm{cls}}^{(t)}$, which is $\ell_2$-normalized. Each class prototype is the
mean of its $K$ support representations, and a query is assigned to the prototype
of highest cosine similarity. Training minimizes the cross-entropy of query
predictions over episodes; the CLIP text encoder is not used.

\subsection{Motion-aligned temporal adapter (MATA)}
\label{sec:mata_adapter}

MATA formulates temporal adaptation (in Fig \ref{fig:overview_and_results1})  in a low-dimensional bottleneck space. 
The key components of MATA are \textit{temporal phase alignment}, \textit{a local motion transition operator}, and residual reconstruction. 
Temporal phase alignment adds frame-order information, the local motion transition operator models short temporal changes, and residual reconstruction maps the temporal response back to the original CLIP token dimension.

\paragraph{Input tokens.}
Let $B$ be the batch size, $T$ the number of frames, $H\times W$ the spatial patch grid, and $C$ the token dimension. 
The token sequence entering a transformer block is
\begin{equation}
X \in \mathbb{R}^{(HW+1)\times (BT)\times C},
\label{eq:mata_input}
\end{equation}
where the first token is the class token and the remaining tokens are spatial patch tokens:
\begin{equation}
X=[x_{\mathrm{cls}};X_p],
\qquad
X_p\in\mathbb{R}^{HW\times(BT)\times C},
\label{eq:mata_token_split}
\end{equation}
so only $X_p$ is processed in the bottleneck space and $x_{\mathrm{cls}}$ remains unchanged. Since the class token attends to patch tokens in the following block, the update
applied to $X_p$ reaches $x_{\mathrm{cls}}$, and hence $z(v)$, through the
remaining blocks.

\begin{table}[t]
\centering

\small
\setlength{\tabcolsep}{2pt}
\renewcommand{\arraystretch}{1.15}

\begin{tabularx}{\columnwidth}{
@{}
l
c
>{\centering\arraybackslash}X
c
c
@{}
}
\toprule
\textbf{Method} &
\textbf{Axis} &
\textbf{Temporal mechanism} &
\textbf{Cost} &
\textbf{Params (M)} \\
\midrule

AIM~\cite{yang2023aim}
& S+T
& Reused temporal MSA
& $O(T^2C)$
& 11.0 \\

ST-Adapter~\cite{pan2022st}
& S+T
& 3D depth-wise conv
& $O(k^3TC)$
& 7.2 \\

DualPath~\cite{park2023dual}
& S+T
& Dual-branch attention
& $O(T^2C)$
& 10.0 \\

D$^2$ST~\cite{pei2025d2st}
& S+T
& Deformable ST attention
& $O(T^2C)$
& 7.3 \\

\midrule

\textbf{MATA}
& \textbf{T}
& \textbf{Depth-wise 1D conv}
& $\boldsymbol{O(kTC_a)}$
& \textbf{3.6} \\

\bottomrule
\end{tabularx}
\caption{Comparison with representative parameter-efficient video adapters.
MATA performs temporal-only adaptation while maintaining linear temporal
complexity. Cost is per patch for the temporal operator alone, at the width at
which each operator acts. $C_a$: bottleneck width; $k$: kernel size; S: spatial;
T: temporal.}
\label{tab:adapter_comparison}
\end{table}


Three properties of the few-shot fine-grained setting motivate this design.
First, CLIP provides strong spatial and semantic priors, and adaptation should
refine them toward temporal discrimination rather than overwrite them, so MATA
updates patch tokens through a residual correction
\begin{equation}
\widehat{X}_p = X_p + \Delta X_p .
\label{eq:mata_residual_update}
\end{equation}
Second, support-query comparison is reliable only when temporally corresponding
action phases are represented consistently, so MATA assigns each frame a learnable phase coordinate before any temporal modeling. Third, fine-grained actions are distinguished by short temporal changes (eg., preparation, onset, release, reversal, delay, or completion), so MATA models transitions among neighboring latent frame states rather than relying on frame-wise appearance alone.

\subsection{MATA Formulation}
\label{sec:mata_formulation}

\paragraph{Bottleneck temporal projection.}
Patch tokens are first projected into a compact adapter space:
\begin{equation}
U=\phi_{\downarrow}(X_p),
\qquad
U\in\mathbb{R}^{HW\times(BT)\times C_a},
\qquad
C_a=\lfloor \alpha C \rfloor ,
\label{eq:mata_down_projection}
\end{equation}
where $\alpha$ is the bottleneck ratio. 
The tensor is then rearranged so that each spatial location forms an independent temporal sequence:
\begin{equation}
U \rightarrow U_f\in\mathbb{R}^{(BHW)\times T\times C_a}.
\label{eq:mata_temporal_reshape}
\end{equation}
This operation isolates temporal evolution while keeping spatial locations structurally separated.

\paragraph{Temporal phase alignment.}
Let $U_f\in\mathbb{R}^{(BHW)\times T\times C_a}$ denote the bottleneck patch sequence, where each index $m$ corresponds to the same spatial patch location observed across frames. 
MATA utilizes a learnable temporal positional embedding
\begin{equation}
P_{\mathrm{time}}\in\mathbb{R}^{T\times C_a}.
\label{eq:mata_phase_embedding}
\end{equation}

The phase-aligned representation is
\begin{equation}
\widetilde{U}_{m,t,c}
=
U_{f,m,t,c}
+
P_{\mathrm{time}}(t,c).
\label{eq:mata_phase_alignment}
\end{equation}

This phase conditioning is achieved through the learnable temporal positional embedding $P_{\mathrm{time}}$, which injects relative frame-order information directly into each patch-level bottleneck feature. Because $P_{\mathrm{time}}$ is indexed by frame position, it is defined for a fixed
$T$ and re-learned when the number of sampled frames changes. 
The \textit{novel design choice is that temporal phase is introduced before local transition modeling}.
Thus, MATA does not treat a patch feature only as a visual state; it represents it as a visual state tied to its relative action phase.

\paragraph{Local motion transition operator.}
After temporal phase alignment, MATA applies temporal specialization along the frame dimension. 
For fine-grained action recognition, this is the relevant adaptation axis: visually similar classes are often separated by short temporal changes rather than by global appearance differences.

For each spatial location $m$, channel $c$, and time index $t$, the temporal response is
\begin{equation}
Y_{m,t,c}
=
\sum_{\delta=0}^{k-1}
w_{c,\delta}
\,
\widetilde{U}^{\mathrm{pad}}_{m,\,t+d\delta-\rho,\,c},
\label{eq:mata_local_transition}
\end{equation}
where $k$ is the temporal kernel size, $d$ is the dilation rate, $w_{c,\delta}$ is the channel-specific temporal weight, and $\rho$ is the padding offset.

\textit{MATA formulates temporal specialization as depth-wise one-dimensional filtering in the bottleneck space, yielding only $kC_a$ temporal parameters.} Per block MATA adds $2CC_a+kC_a+TC_a$ parameters; with $C{=}768$, $\alpha{=}0.25$,
$k{=}3$, $T{=}8$ over 12 blocks this gives the $3.6$M reported in
Table~\ref{tab:adapter_comparison}.
The operation is also compute-efficient because it acts locally along $T$ frames at fixed spatial locations, avoiding dense temporal attention over all video tokens.

\paragraph{Residual reconstruction.}
The temporal response is mapped back to the original token dimension and added to the input patch tokens:
\begin{equation}
\Delta X_p
=
\phi_{\uparrow}(\sigma(Y)),
\qquad
\widehat{X}
=
[x_{\mathrm{cls}};\,X_p+\Delta X_p].
\label{eq:mata_reconstruction}
\end{equation}
The up-projection $\phi_{\uparrow}$ is zero-initialized, so the adapter starts as an identity mapping and gradually learns motion-specific corrections.

\paragraph{Compact form.}
The complete transformation can be summarized as
\begin{equation}
\widehat{X}
=
\left[
x_{\mathrm{cls}};
\,
X_p+
\phi_{\uparrow}
\left(
\sigma
\left(
F_{\mathrm{temp}}
\left(
\phi_{\downarrow}(X_p)+P_{\mathrm{time}}
\right)
\right)
\right)
\right].
\label{eq:mata_compact}
\end{equation}
Thus, MATA consists of three coupled operations: temporal phase alignment, local motion transition operator, and residual reconstruction.

Table~\ref{tab:adapter_comparison} compares MATA among representative
parameter-efficient video adapters. Where prior adapters jointly model
spatial appearance and temporal structure through multi-branch,
3D-convolutional, or attention-based modules, MATA adapts \emph{only} the
temporal axis, and assigns each frame a learnable temporal phase \emph{before}
modeling local transitions, an ordering that ties each patch feature to its
relative action phase and that we isolate in Table~\ref{tab:tconv_pe_study}. Acting
in the bottleneck space, its temporal cost is linear rather than quadratic in
$T$. MATA thus trades generality for focus, which we show is advantageous when
the discriminative signal is short-lived motion rather than appearance.

\begin{table*}[t]

\centering
\small
\setlength{\tabcolsep}{3pt}
\begin{tabular}{l l l | ccc | ccc | ccc}
\toprule
\textbf{Method} & \textbf{Venue} & \textbf{\shortstack{Tunable\\Params (M)}} &
\multicolumn{3}{c|}{\textbf{Diving48}} &
\multicolumn{3}{c|}{\textbf{FineGym99}} &
\multicolumn{3}{c}{\textbf{FineGym288}} \\
\cmidrule(lr){4-6}
\cmidrule(lr){7-9}
\cmidrule(lr){10-12}
& & &
1-shot & 3-shot & 5-shot &
1-shot & 3-shot & 5-shot &
1-shot & 3-shot & 5-shot \\
\midrule

TSN+Cosine \cite{wang2016temporal}& ECCV'16 & 25M &
29.72 & 35.69 & 38.81 &
36.63 & 41.35 & 42.76 &
36.68 & 42.87 & 44.82 \\

ProtoNet \cite{snell2017prototypical} & NeurIPS'17 & 25M &
59.80 & 74.74 & 77.69 &
62.99 & 74.25 & 77.99 &
58.98 & 70.56 & 74.87 \\

OTAM \cite{cao2020few} &  CVPR'20 & 23.5M &
41.64 & 47.96 & 49.78  &
48.93 & 52.88 & 55.61  &
47.58 & 53.67 & 55.62  \\

PAL \cite{zhu2021few}& BMVC’21 & 25M &
50.40  & 57.57 & 61.17 &
58.85  & 65.13 & 68.64 &
58.32  & 63.62 & 65.77 \\

TRX \cite{perrett2021temporal}& CVPR'21 & 47.1M &
62.53  & 77.81 & 82.01 &
66.55  & 80.15 & 83.86 &
64.03  & 76.73 & 81.05 \\

STRM \cite{thatipelli2022spatio} &  CVPR'22 & 30M &
62.29 & 77.79 & 81.03 &
68.09 & 80.93 & 84.33 &
64.81 & 76.98 & 80.64 \\

HyRSM \cite{wang2022hybrid} &  CVPR'22 & 65.6M &
61.04 & 76.04 & 81.74 &
67.09 & 78.15 & 82.92 &
62.86 & 73.12 & 78.81 \\

MoLo \cite{wang2023molo} &  CVPR'23 & 89.6M &
- & - & - &
73.3 & 80.2 & 84.8 &
- & - & - \\

TATs \cite{kumar2024trajectory} &  ECCV'24 & 10.8M &
- & - & - &
81.8 & 86.0 & 87.9 &
- & - & - \\

M$^3$Net \cite{tang2023m3net}&  MM'23 & 30M &
72.35 & 82.50 & 85.44 &
72.70 & 83.03 & 86.71 &
68.66 & 79.73 & 83.05 \\

\midrule
\multicolumn{12}{c}{\textit{DINOv2 Backbone}} \\
\midrule
Trokens\cite{kumar2025trokens}$^{\phi}$ &  ICCV'25 & 17.4M &
67.20 $^{\dagger}$  & 77.14 $^{\dagger}$  & 81.30 $^{\dagger}$  &
\textbf{84.40} & 88.30 & 89.90 &
- & - & - \\

\midrule
\multicolumn{12}{c}{\textit{CLIP Backbone}} \\
\midrule

D$^2$ST \cite{pei2025d2st}$^{\dagger}$ & ICCV'25 & 7.3M &
73.29 & 81.64 & 83.85 &
76.81 & 85.52 & 87.55 &
70.33 & 76.26 & 82.08 \\

EMP-Net \cite{wu2024efficient}$^{\dagger}$* &  ECCV'24 & 7.5M &
71.00 & - & 69.20 &
- & - & - &
- & - & - \\

Task-Adapter \cite{cao2024task}$^{\dagger}$* &  MM'24 & 7.2M &
68.04 & 72.47 & 75.09 &
64.48 & 67.12 & 70.54 &
- & - & - \\

Task-Adapter++ \cite{cao2026task}$^{\dagger}$* &  PR'26 & 8.0M &
46.45 & 60.65 & 64.17 &
60.82 & 65.43 & 68.12 &
63.07 & - & 72.76 \\

TEAM \cite{lee2025temporal}$^{\dagger}$ &  CVPR'25 & 87.0M & 38.50 & 50.40 & 54.90 &
- & - & - &
- & - & - \\

MPL \cite{yang2026mpl}$^{\dagger}$ &  CVPR'26 & 14.0M &
39.29 & 44.61 & 48.7 &
- & - & - &
- & - & - \\

\midrule

\rowcolor{green!12} \textbf{MATA} & - & \textbf{3.6M} &
\textbf{\shortstack{80.12 \\ {\scriptsize $\pm$ 0.15}}} & 
\textbf{\shortstack{88.31 \\ {\scriptsize $\pm$ 0.14}}} & 
\textbf{\shortstack{90.00 \\ {\scriptsize $\pm$ 0.12}}} & 
\shortstack{82.21 \\ {\scriptsize $\pm$ 0.18}} & 
\textbf{\shortstack{88.97 \\ {\scriptsize $\pm$ 0.16}}} & 
\textbf{\shortstack{91.29 \\ {\scriptsize $\pm$ 0.13}}} &
\textbf{\shortstack{81.28 \\ {\scriptsize $\pm$ 0.17}}} & 
\textbf{\shortstack{89.39 \\ {\scriptsize $\pm$ 0.15}}} & 
\textbf{\shortstack{90.13 \\ {\scriptsize $\pm$ 0.19}}} \\

\bottomrule
\end{tabular}
\caption{\textbf{Few-shot fine-grained action recognition on Diving48, FineGym99, and FineGym288}  (\%)  under 5-way 1/3/5-shot settings. This table compares prior few-shot methods that do not use foundation models with MATA. Best results are in bold.\textit{$^{\dagger}$ means results reproduced by us.  “*” denotes multi-modal methods incorporating additional text input and a text encoder. $\phi$ denotes additional pre-processing.}}
\label{tab:maincomparison}
\end{table*}


\begin{table*}[t]

\centering
\small
\setlength{\tabcolsep}{3pt}

\begin{tabular}{l|ccc|ccc|ccc}
\toprule
\textbf{Method}
& \multicolumn{3}{c|}{\textbf{ToyotaSH}}
& \multicolumn{3}{c|}{\textbf{HAA500}}
& \multicolumn{3}{c}{\textbf{MPII Cooking 2}} \\
\cmidrule(lr){2-4}
\cmidrule(lr){5-7}
\cmidrule(lr){8-10}
& 1-shot & 3-shot & 5-shot
& 1-shot & 3-shot & 5-shot
& 1-shot & 3-shot & 5-shot \\
\midrule

Task-Adapter++ \cite{cao2026task}$^{\dagger}$
& 62.39 & 73.43 & 76.30
& 97.05 & 98.14 & 99.10
& 59.85 & 68.79 & 70.00 \\

D$^2$ST \cite{pei2025d2st}$^{\dagger}$
& 60.25 & 71.06 & 73.18
& 96.52 & 97.46 & 98.78
& 57.23 & 68.09 & 71.16 \\

\midrule
\rowcolor{green!12} MATA
& \textbf{64.84} & \textbf{74.85} & \textbf{77.76}
& \textbf{97.18} & \textbf{98.24} & \textbf{99.28}
& \textbf{60.03} & \textbf{70.48} & \textbf{72.50} \\

\bottomrule
\end{tabular}
\caption{\textbf{Few-shot fine-grained action recognition on additional datasets} (\%) under 5-way 1/3/5-shot settings.}
\label{tab:fewshot_comparison_other_ds}
\end{table*}

\begin{table*}[t]

\small

\resizebox{0.9\linewidth}{!}{%
\begin{tabular}{lccccc}
\midrule
Method &
\shortstack{1-shot Acc.\\(\%) $\uparrow$} &
\shortstack{GFLOPs\\Episode $\downarrow$} &
\shortstack{Latency\\Episode (ms) $\downarrow$} &
\shortstack{Throughput\\(episodes/s) $\uparrow$} &
\shortstack{Effective Compute\\(TFLOP/s) $\uparrow$} \\
\midrule
D$^2$ST $^{\dagger}$ \cite{pei2025d2st} & 73.29 & 2990.0 & 111.26 & 9.07  & 26.92  \\
Task-Adapter $^{\dagger}$ \cite{cao2026task} & 46.45 & 3335.95 & 68.54 & 14.60  & 48.67 \\
EMP-Net $^{\dagger}$ \cite{wu2024efficient} & 71.0 & 2964.24 & 128.16 & 7.80  & 23.12  \\
MATA    & \textbf{80.12} & \textbf{2921.3} & \textbf{27.16}  &\textbf{ 36.81} & \textbf{107.53}  \\
\midrule
\rowcolor{green!12} Gain (MATA vs. D$^2$ST) &
$\uparrow$ +6.83 &
$\downarrow$ 2.3\% &
$\downarrow$ 4.10$\times$ &
$\uparrow$ 4.06$\times$ &
$\uparrow$ 3.99$\times$ \\
\midrule
\end{tabular}%
}
\centering
\caption{\textbf{MATA delivers both higher accuracy and substantially faster few-shot adaptation.} The results reveal a clear advantage in both recognition performance and practical inference efficiency, rather than a trade-off between the two. $\dagger$ denotes results reproduced by us; $\downarrow$ lower is better and $\uparrow$ higher is better.}
\label{tab:fewshot_efficiency_mata_d2st}
\end{table*}

\section{Results and discussion}
\label{sec:exp}

\subsection{Implementation details}
We evaluate MATA on six challenging few-shot action benchmarks: Diving48~\citep{li2018diving48}, FineGym99, FineGym288~\citep{shao2020finegym}, Toyota SmartHome~\citep{das2019toyota}, HAA500~\citep{chung2021haa500} and MPII Cooking2 \cite{rohrbach15ijcv}. These datasets cover fine-grained sports, daily activities, human-object interactions, and temporally sensitive action classes. We use the same class splits as M$^3$Net~\citep{tang2023m3net} where applicable for fair comparison.

We follow a 5-way episodic meta-learning protocol with 60{,}000 training episodes. Only the MATA adapter is optimized with query loss; the CLIP backbone remains frozen. Unless otherwise stated, each video is represented by 8 segment-sampled frames, and training uses batch size of 8, standard video augmentations, and NVIDIA H100 GPUs. At test time, we evaluate 10{,}000 episodes under 1-shot, 3-shot, and 5-shot settings and report Top-1 accuracy with single-view inference. All experiments use a frozen CLIP ViT-B/16 backbone and 8 input frames by default. MATA uses temporal kernel size $k{=}3$, dilation $d{=}2$, and bottleneck ratio $0.25$.


\subsection{Establishing state-of-the-art performance across benchmarks}
\label{sec:perf_comparison}

MATA is the most accurate and the most parameter-efficient method in Table~\ref{tab:maincomparison} in nearly all settings. It improves over the strongest non-foundation baseline by up to $+12.62$ points at 1-shot with $90\%$ fewer parameters than M$^3$Net, and over D$^2$ST, the strongest prior adapter, by up to $+10.95$ points with half the tunable parameters. Gains are largest in the low-shot regime, where support-query temporal alignment matters most. Trokens is the closest competitor, but on unequal terms: it uses a stronger DINOv2 backbone and trajectory tokens built in an \textit{offline stage before training}, paying in \textit{preprocessing cost}, \textit{storage}, and \textit{sensitivity to tracker quality}. It leads on FineGym99 at 1-shot; MATA leads at 3- and 5-shot and Diving48 by $+12.92$ points, with $79\%$ fewer parameters and a single pass over raw frames. Gains extend beyond sports motion. Table~\ref{tab:fewshot_comparison_other_ds} shows consistent improvements on Toyota SmartHome, HAA500, and MPII Cooking2, where the cues are object manipulation and interaction order rather than body dynamics. Coarse-grained comparisons are in Sec.~\ref{sec:coarse_results} of the supplementary.

\subsection{Achieving the accuracy--compute efficiency--unimodality objective}
\label{sec:fewshot_efficiency_mata_d2st}

The accuracy--compute efficiency--unimodality objective asks whether a few-shot video model can improve recognition while remaining compact, fast, and single-stream. Therefore, we compare episodic inference efficiency between MATA and D$^2$ST under an identical few-shot protocol on a single NVIDIA H100 80GB HBM3 GPU using FP16. FLOPs are measured with \texttt{fvcore}, and latency is measured with CUDA events over timed blocks of $B{=}8$ consecutive meta-forwards. Each episode uses a 5-way 1-shot setting with one query per class and $T{=}8$ frames, corresponding to 40 support and 40 query frame-rows per episode.

Table~\ref{tab:fewshot_efficiency_mata_d2st} shows that MATA improves both accuracy and runtime efficiency under the same episodic protocol. At comparable nominal FLOPs, MATA is more accurate and cuts per-episode latency from $111.26$ to $27.16$ ms, a $4.10\times$ speedup that raises throughput from $9.07$ to $36.81$ episodes/s and realized compute from $26.92$ to $107.53$ TFLOP/s.

FLOPs therefore do not predict few-shot inference cost. D$^2$ST's deformable attention issues many small, irregular memory accesses, keeping GPU utilization low regardless of arithmetic volume ($26.92$ TFLOP/s realized). MATA's temporal operator is a depth-wise convolution over a contiguous frame axis, which maps to dense kernels and leaves the frozen CLIP pathway untouched ($107.53$ TFLOP/s). The gain comes from what MATA does not add to that pathway, not from doing less arithmetic.

\subsection{MATA is scalable and applicable to a fully-supervised setting.}
MATA is not restricted to the few-shot regime. Under the standard fully-supervised protocol, it reaches $89.7\%$ on Diving48 (Table~\ref{tab:diving48_sota}) and $95.1\%$ / $90.0\%$ on FineGym99 and FineGym288 (Table~\ref{tab:finegym_sota}), matching or exceeding AIM and DualPath with roughly a third of their tunable parameters and outperforming backbones an order of magnitude larger. MATA also scales with backbone capacity: Table~\ref{tab:mata_backbone_comparison} shows consistent gains from ViT-B/16 to ViT-L/14 across all three fine-grained benchmarks, indicating that temporal-only adaptation continues to benefit from stronger spatial priors rather than saturating.

\begin{table*}[t]
\centering
\small

\begin{minipage}[t]{0.47\textwidth}
\vspace{0pt}
\centering

\resizebox{\linewidth}{!}{%
\begin{tabular}{lcccc}
\toprule
\textbf{Method} & \textbf{Venue} &
\textbf{\shortstack{Tunable\\Params (M)}} &
\textbf{Frames} &
\textbf{Top-1} \\
\midrule
\multicolumn{5}{c}{\textit{Non-CLIP-based methods}} \\
\midrule
ORViT TimeSformer \cite{herzig2022object}
    & CVPR'22 & 160.0 & 32 & 88.0 \\
BEVT \cite{wang2022bevt}
    & CVPR'22 & 88.1 & 16 & 86.7 \\
VideoSwin-B \cite{liu2022video}
    & CVPR'22 & 88.1 & 32 & 81.9 \\
SlowFast R101 \cite{feichtenhofer2019slowfast}
    & ICCV'19 & 62.8 & 64+16 & 77.6 \\
TQN \cite{zhang2021temporal}
    & CVPR'21 & -- & all & 81.8 \\
\midrule
\multicolumn{5}{c}{\textit{CLIP-based methods ViT-B/16}} \\
\midrule
BDC-CLIP $^{\dagger}$ \cite{long2025bdc}
    & ICML'25 & 125.4 & 16 & 76.6 \\
PeVL \cite{zhang2024pevl}
    & CVPR'24 & 28.0 & 16 & 84.5 \\
DiST $^{\dagger}$ \cite{qing2023disentangling}
    & ICCV'23 & 19.0 & 16 & 76.85 \\
AIM $^{\dagger}$ \cite{yang2023aim}
    & ICLR'23 & 11.0 & 32 & 88.9 \\
DualPath $^{\dagger}$ \cite{park2023dual}
    & CVPR'23 & 10.0 & 32 & 88.7 \\
ST Adapter $^{\dagger}$ \cite{pan2022st}
    & NeurIPS'22 & 7.2 & 16 & 84.2 \\
\midrule
MATA (ours)
    & -- & 3.6 & 16 & 85.8 $\pm$ 0.13 \\
\rowcolor{green!12}
MATA (ours)
    & -- & 3.6 & 32 & \textbf{89.7 $\pm$ 0.15} \\
\bottomrule
\end{tabular}%
}

\caption{\textbf{Fully-supervised results on Diving48.}
Top-1 accuracy under the standard fully-supervised protocol,
included to show MATA also applies beyond the few-shot setting.}
\label{tab:diving48_sota}
\end{minipage}
\hfill
\begin{minipage}[t]{0.51\textwidth}
\vspace{0pt}
\centering

\resizebox{\linewidth}{!}{%
\begin{tabular}{lccccc}
\toprule
\textbf{Method} & \textbf{Venue} &
\textbf{Frames} &
\textbf{\shortstack{Tunable\\Params (M)}} &
\multicolumn{2}{c}{\textbf{Top-1}} \\
\cmidrule(lr){5-6}
& & & & \textbf{FineGym99} & \textbf{FineGym288} \\
\midrule
\multicolumn{6}{c}{\textit{Non-CLIP-based methods}} \\
\midrule
TQN \cite{zhang2021temporal}
    & CVPR'21 & all & 78.0 & 93.8 & 89.6 \\
TSM \cite{lin2019tsm}
    & ICCV'19 & 16+16 & 48.6 & 88.4 & 83.1 \\
TRNms \cite{zhou2018temporal}
    & ECCV'18 & 8+8 & 36.6 & 87.8 & 82.0 \\
SlowOnly \cite{feichtenhofer2019slowfast}
    & ICCV'19 & 32 & 32.4 & 93.46 & 86.8 \\
I3D \cite{ao2017quo}
    & CVPR'17 & 16 & 28.0 & 75.6 & 77.3 \\
\midrule
\multicolumn{6}{c}{\textit{CLIP-based methods ViT-B/16}} \\
\midrule
PeVL$^{\dagger}$ \cite{zhang2024pevl}
    & CVPR'24 & 16 & 28.0 & 91.2 & 85.4 \\
AIM $^{\dagger}$ \cite{yang2023aim}
    & ICLR'23 & 32 & 11.0 & 94.9 & 89.1 \\
\midrule
\rowcolor{green!12}
MATA (ours)
    & -- & 32 & 3.7
    & \textbf{95.1 $\pm$ 0.13}
    & \textbf{90.0 $\pm$ 0.10} \\
\bottomrule
\end{tabular}%
}

\caption{\textbf{Fully-supervised results on FineGym99 and
FineGym288.} PeVL \cite{zhang2024pevl} results are reported
using video modality.}
\label{tab:finegym_sota}
\end{minipage}

\par\vspace{3mm}

\begin{minipage}{0.75\textwidth}
\centering
\small
\setlength{\tabcolsep}{3pt}

\begin{tabular}{l|c|cc|cc|cc}
\toprule
\multirow{2}{*}{MATA} &
\multirow{2}{*}{Params} &
\multicolumn{2}{c|}{Diving48} &
\multicolumn{2}{c|}{FineGym99} &
\multicolumn{2}{c}{FineGym288} \\
\cmidrule(lr){3-4}
\cmidrule(lr){5-6}
\cmidrule(lr){7-8}
& & 1-shot & 5-shot
  & 1-shot & 5-shot
  & 1-shot & 5-shot \\
\midrule
ViT-B/16
& 3.6
& 80.12 & 90.00
& 82.21 & 91.29
& 81.28 & 90.13 \\
ViT-L/14
& 12.5
& \textbf{82.65} & \textbf{90.00}
& \textbf{83.07} & \textbf{92.53}
& \textbf{83.00} & \textbf{91.64} \\
\bottomrule
\end{tabular}

\caption{\textbf{Backbone scalability of MATA.}
Top-1 accuracy (\%) under 1-shot and 5-shot settings with
frozen CLIP backbones. Gains carry over to the larger
backbone across all three benchmarks.}
\label{tab:mata_backbone_comparison}
\end{minipage}

\vspace{-6mm}
\end{table*}

\subsection{Ablation study}

We ablate the core components of MATA to isolate the role of temporal phase alignment. Table~\ref{tab:tconv_pe_study} shows that adding temporal phase alignment to the local motion transition operator consistently improves both 1-shot and 5-shot accuracy on Diving48 and FineGym99. This indicates that local motion transition modeling alone is not sufficient: fine-grained support-query matching also benefits from knowing where a motion state occurs within the action sequence. Additional ablations on adapter depth, input frames, and supervision are provided in Sec.~\ref{sec:scaling} in the supplementary.

\begin{table}[t]
\centering
\small
\setlength{\tabcolsep}{5pt}

\begin{tabular}{l|cc|cc}
\toprule
\multirow{2}{*}{Component} &
\multicolumn{2}{c|}{Diving48} &
\multicolumn{2}{c}{FineGym99} \\
\cmidrule(lr){2-3}
\cmidrule(lr){4-5}
& 1-shot & 5-shot & 1-shot & 5-shot \\
\midrule
Local motion transition
    & 78.69 & 88.76 & 80.02 & 89.26 \\
\quad + Temporal phase alignment
    & \textbf{80.12} & \textbf{90.00}
    & \textbf{82.21} & \textbf{91.29} \\
\bottomrule
\end{tabular}

\caption{\textbf{Component analysis} of MATA on Diving48 and FineGym99.
Each row cumulatively adds the indicated component.}
\label{tab:tconv_pe_study}
\vspace{-6mm}
\end{table}

\subsection{MATA remains robust against distribution shifts.}
We evaluate 5-way 1-shot recognition under five test-time corruptions at
severity levels $\{1,3,5\}$. Support videos remain clean and only query clips
are corrupted, isolating robustness to degraded test inputs. We consider
Gaussian noise, Gaussian blur, JPEG compression, brightness, and frame drop;
the latter removes 12.5\%/37.5\%/62.5\% of the 8 frames and fills missing
positions by hold-last, providing a temporal stress test. Corruptions are
applied in pixel space before CLIP normalization, and each method is evaluated
on 2,000 episodes.

We report mean Corruption Accuracy (mCA) and mean Corruption Error (mCE). Following the mCE metric introduced with ImageNet-C~\cite{hendrycks2019benchmarking}, we adapt mCE to our video robustness setting. Let $\bar{A}_{m,c}$ denote the Top-1 accuracy of method $m$ on corruption $c$, averaged over severities $\{1,3,5\}$. We define
{\setlength{\abovedisplayskip}{3pt}
 \setlength{\belowdisplayskip}{3pt}
 \setlength{\abovedisplayshortskip}{2pt}
 \setlength{\belowdisplayshortskip}{2pt}
\begin{equation}
\mathrm{mCE}_m = 20\sum_{c=1}^{5}
\left(100-\bar{A}_{m,c}\right)\big/\left(100-\bar{A}_{\mathrm{ref},c}\right),
\label{eq:mce}
\end{equation}}
\noindent where Task-Adapter++ is the reference method. Thus, mCA measures absolute corrupted accuracy, while mCE measures corruption error relative to a common reference, with lower mCE indicating greater robustness.

Table~\ref{tab:robustness} shows that MATA achieves the highest accuracy under every corruption, improving mCA over D$^2$ST by 8.0 points (72.8 vs.\ 64.8). MATA also obtains the lowest mCE, reducing normalized corruption error from 57.5 for D$^2$ST to 44.3, a 13.2-point improvement. These results show that MATA's gains are not limited to clean inputs: its temporal specialization
remains effective under both spatial corruptions and temporal degradation.

\begin{table*}[h]
\centering
\setlength{\tabcolsep}{3.5pt}
\begin{tabular}{lcccccccc}
\toprule
\multirow{2}{*}{Method} &
\multirow{2}{*}{\shortstack{Clean\\Top-1 $\uparrow$}} &
\multicolumn{5}{c}{Corrupted queries (mean Top-1 over severities 1/3/5) $\uparrow$} &
\multirow{2}{*}{\shortstack{mCA\\(\%) $\uparrow$}} &
\multirow{2}{*}{\shortstack{mCE\\$\downarrow$}} \\
\cmidrule(lr){3-7}
& &
\shortstack{Gaussian\\noise} &
\shortstack{Gaussian\\blur} &
\shortstack{Frame\\drop} &
\shortstack{JPEG\\compr.} &
\shortstack{Bright-\\ness} & & \\
\midrule
Task-Adapter++ & 45.8 & 32.1 & 36.6 & 44.1 & 40.8 & 41.0 & 38.9 & 100.0 \\
D$^2$ST        & 73.8 & 55.4 & 62.9 & 66.9 & 68.5 & 70.1 & 64.8 & 57.5 \\
\midrule
\rowcolor{green!12}
MATA & \textbf{80.1} & \textbf{63.7} & \textbf{70.7} &
\textbf{74.2} & \textbf{77.4} & \textbf{77.9} &
\textbf{72.8} & \textbf{44.3} \\
\bottomrule
\end{tabular}
\caption{\textbf{Test-time robustness} on Diving48 (5-way 1-shot); only queries are corrupted. Mean Top-1 (\%) over severities $\{1,3,5\}$; mCE is normalized by Task-Adapter++ (=100, lower is better).}
\label{tab:robustness}
\end{table*}

\subsection{Why MATA's temporal modeling captures fine-grained details}

We provide qualitative and statistical analyses to explain why MATA consistently outperforms D$^2$ST. We further include a \textit{single-frame blackout sensitivity analysis} in Sec. \ref{sec:blackout} to measure temporal selectivity under targeted frame removal and nearest-neighbor retrieval across confusing classes in Sec. \ref{sec:nn-retrieval} in the supplementary.

\subsubsection{Qualitative diagnostic on temporal saliency}
\label{sec:saliency}

\begin{figure}[h]
  \centering
  \includegraphics[width=\linewidth]{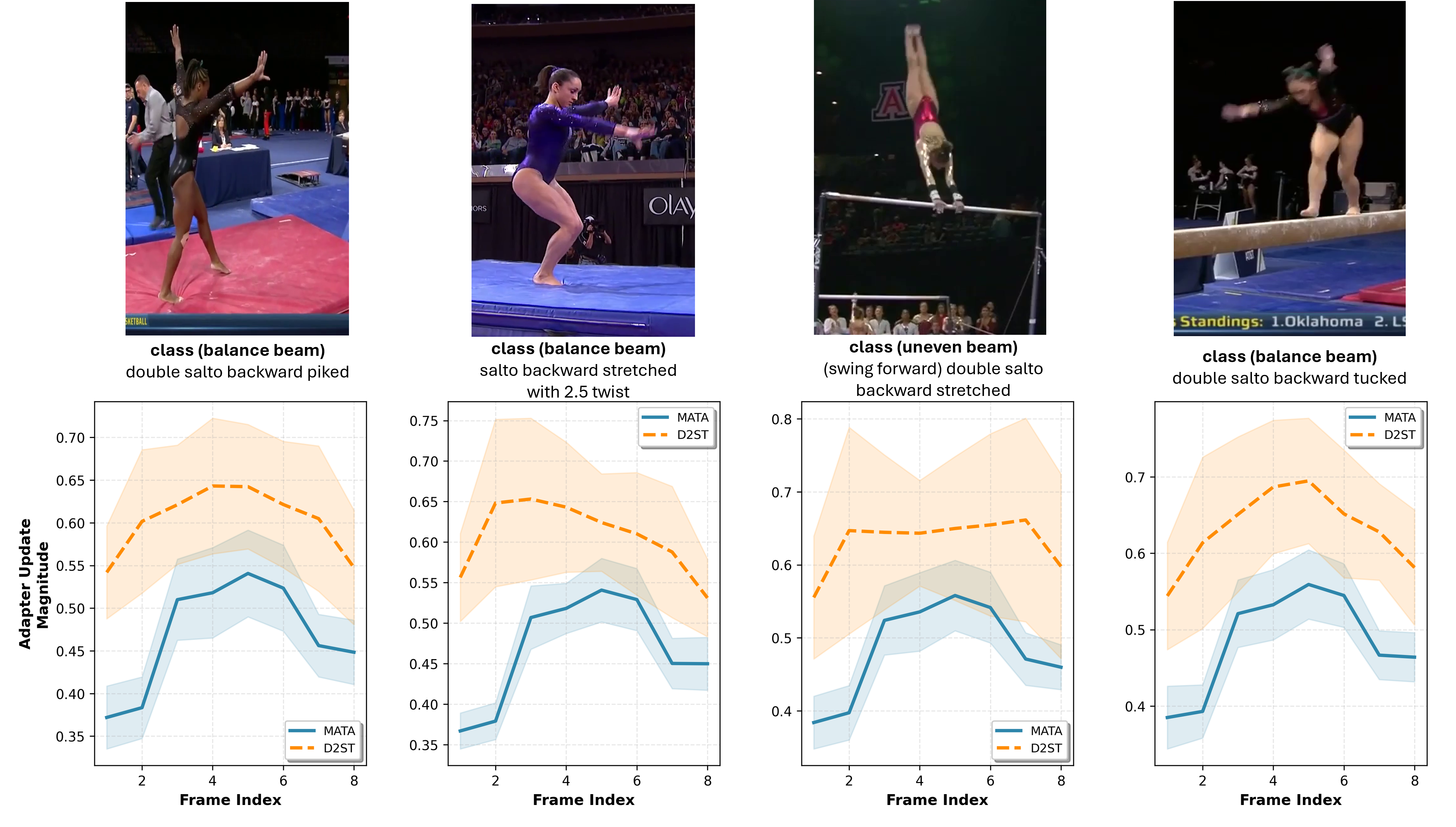}
  \caption{\textbf{Temporal saliency via adapter update magnitude}: Per-frame update magnitude $s(t)$ (Eq.~\ref{eq:update_mag}) for MATA and D$^2$ST on randomly sampled FineGym clips (mean $\pm$ variability). MATA shows sharper temporal peaks, indicating more selective activation on discriminative motion stages. More comparison examples in Sec. \ref{sec:tempmore} in the supplementary.}
  \label{fig:magnitude1}
\end{figure}

We analyze \emph{temporal saliency} to explain why MATA improves fine-grained discrimination through temporally localized adaptation. Fine-grained actions often differ by brief stage transitions and motion ordering, so effective adaptation should be temporally selective rather than uniformly modifying representations across the clip. MATA explicitly biases adaptation toward local temporal cues via a depth-wise 1D temporal convolution and learned temporal positions, aiming to spend limited capacity on the most discriminative frames.

We measure temporal saliency using the per-frame magnitude of the adapter-induced residual update. Let $\Delta X_{\text{patch}} \in \mathbb{R}^{HW \times (BT) \times C}$ denote the adapter residual update in a transformer block (Sec \ref{sec:method}). Reshaping to isolate time, we define
\begin{equation}
s(t) \;=\; \frac{1}{HW}\sum_{n=1}^{HW} \left\lVert \Delta X_{n,t} \right\rVert_2,
\label{eq:update_mag}
\end{equation}
where $n$ indexes spatial tokens and $t \in \{1,\dots,T\}$ indexes frames. Intuitively, $s(t)$ quantifies how strongly the adapter modifies the representation at time $t$. We report the mean curve with variability (shaded region) across multiple episodes on randomly sampled FineGym classes, aggregating across examples to reflect a generalized trend.

\noindent\textbf{Selective peaks vs.\ uniform updates.}
Fig.~\ref{fig:magnitude1} shows that MATA yields a lower baseline update magnitude with sharp temporal peaks concentrated around discriminative stages, consistent with fine-grained cues being short-lived. In contrast, D$^2$ST~\cite{pei2025d2st} exhibits a higher and flatter response, suggesting more uniform perturbations across frames. These saliency patterns support the intended mechanism of MATA: it applies small, targeted modifications and activates strongly only near key transition moments, reducing unnecessary feature drift and thereby aiding few-shot generalization.

\section{Conclusion and limitations}
\label{sec:conclusion}

We presented MATA, a motion-aligned temporal adapter for few-shot fine-grained action recognition with a frozen CLIP backbone. By combining temporal phase alignment with local modeling of short motion transitions, MATA achieves state-of-the-art results on fine-grained benchmarks with only $\sim$3.6M tunable parameters, lower latency, and the best accuracy under every test-time corruption, without full backbone tuning or auxiliary modalities. MATA is specialized by design: it assumes a fixed frame count and does not target long-range or appearance-driven recognition, leaving long-duration benchmarks and streaming recognition for future work.

\noindent\textbf{Acknowledgment}: This work has emanated from research supported by the Kempe Foundations (Kempestiftelserna), JCSMK23-0109, Sweden.

{
    \small
    \bibliographystyle{ieeenat_fullname}
    \bibliography{example_paper}

@inproceedings{xu2021actionclip,
  title={{ActionCLIP}: a new paradigm for video action recognition},
  author={Xu, Mengmeng and Zhang, Han and Xu, Jian and others},
  booktitle={Proceedings of the IEEE/CVF International Conference on Computer Vision (ICCV)},
  year={2021}
}

@inproceedings{pei2025d2st,
  title={{D2ST}-Adapter: disentangled and deformable spatio-temporal adapter for few-shot action recognition},
  author={Pei, Wenjie and Tan, Qizhong and Lu, Guangming and Tian, Jiandong and Yu, Jun},
  booktitle={Proceedings of the IEEE/CVF International Conference on Computer Vision (ICCV)},
  pages={11317--11326},
  year={2025}
}

@inproceedings{yang2023aim,
  title={{AIM}: adapting image models for efficient video action recognition},
  author={Yang, Taojiannan and Zhu, Yi and Xie, Yusheng and Zhang, Aston and Chen, Chen and Li, Mu},
  booktitle={International Conference on Learning Representations (ICLR)},
  year={2023}
}

@inproceedings{li2018diving48,
  title={{RESOUND}: towards action recognition without representation bias},
  author={Li, Yingwei and Li, Yi and Vasconcelos, Nuno},
  booktitle={Proceedings of the European Conference on Computer Vision (ECCV)},
  pages={513--528},
  year={2018}
}

@inproceedings{shao2020finegym,
  title={{FineGym}: a hierarchical video dataset for fine-grained action understanding},
  author={Shao, Dian and Zhao, Yue and Dai, Bo and Lin, Dahua},
  booktitle={Proceedings of the IEEE/CVF Conference on Computer Vision and Pattern Recognition (CVPR)},
  pages={2616--2625},
  year={2020}
}

@inproceedings{zhang2024pevl,
  title={{PEVL}: pose-enhanced vision-language model for fine-grained human action recognition},
  author={Zhang, Haosong and Leong, Mei Chee and Li, Liyuan and Lin, Weisi},
  booktitle={Proceedings of the IEEE/CVF Conference on Computer Vision and Pattern Recognition (CVPR)},
  pages={18857--18867},
  year={2024}
}

@inproceedings{tang2023m3net,
  title={{M3Net}: Multi-View Encoding, Matching, and Fusion for Few-Shot Fine-Grained Action Recognition},
  author={Tang, Hao and Liu, Jun and Yan, Shuanglin and Yan, Rui and Li, Zechao and Tang, Jinhui},
  booktitle={Proceedings of the ACM International Conference on Multimedia (ACM MM)},
  pages={1719--1728},
  year={2023}
}

@article{wang2024clip,
  title={{CLIP}-Guided Prototype Modulating for Few-Shot Action Recognition},
  author={Wang, Xiang and Zhang, Shiwei and Cen, Jun and Gao, Changxin and Zhang, Yingya and Zhao, Deli and Sang, Nong},
  journal={International Journal of Computer Vision},
  volume={132},
  number={6},
  pages={1899--1912},
  year={2024}
}

@inproceedings{houlsby2019parameter,
  title={Parameter-efficient transfer learning for NLP},
  author={Houlsby, Neil and Giurgiu, Andrei and Jastrzebski, Stanislaw and Morrone, Bruna and De Laroussilhe, Quentin and Gesmundo, Andrea and Attariyan, Mona and Gelly, Sylvain},
  booktitle={Proceedings of the International Conference on Machine Learning (ICML)},
  pages={2790--2799},
  year={2019}
}

@inproceedings{das2019toyota,
  title={Toyota smarthome: Real-world activities of daily living},
  author={Das, Srijan and Dai, Rui and Koperski, Michal and Minciullo, Luca and Garattoni, Lorenzo and Bremond, Francois and Francesca, Gianpiero},
  booktitle={Proceedings of the IEEE/CVF Conference on Computer Vision and Pattern Recognition (CVPR)},
  pages={833--842},
  year={2019}
}

@inproceedings{radford2021clip,
  title={Learning transferable visual models from natural language supervision},
  author={Radford, Alec and Kim, Jong Wook and Hallacy, Chris and Ramesh, Aditya and Goh, Gabriel and Agarwal, Sandhini and Sastry, Girish and Askell, Amanda and Mishkin, Pamela and Clark, Jack and others},
  booktitle={Proceedings of the International Conference on Machine Learning (ICML)},
  pages={8748--8763},
  year={2021}
}

@article{pan2022st,
  title={{ST}-Adapter: parameter-efficient image-to-video transfer learning},
  author={Pan, Junting and Lin, Ziyi and Zhu, Xiatian and Shao, Jing and Li, Hongsheng},
  journal={Advances in Neural Information Processing Systems},
  volume={35},
  pages={26462--26477},
  year={2022}
}

@inproceedings{chung2021haa500,
  title={{HAA500}: human-centric atomic action dataset with curated videos},
  author={Chung, Jihoon and Wuu, Cheng-hsin and Yang, Hsuan-ru and Tai, Yu-Wing and Tang, Chi-Keung},
  booktitle={Proceedings of the IEEE/CVF Conference on Computer Vision and Pattern Recognition (CVPR)},
  pages={13465--13474},
  year={2021}
}

@inproceedings{park2023dual,
  title={Dual-path adaptation from image to video transformers},
  author={Park, Jungin and Lee, Jiyoung and Sohn, Kwanghoon},
  booktitle={Proceedings of the IEEE/CVF Conference on Computer Vision and Pattern Recognition (CVPR)},
  pages={2203--2213},
  year={2023}
}

@inproceedings{rasheed2023fine,
  title={Fine-tuned {CLIP} models are efficient video learners},
  author={Rasheed, Hanoona and Khattak, Muhammad Uzair and Maaz, Muhammad and Khan, Salman and Khan, Fahad Shahbaz},
  booktitle={Proceedings of the IEEE/CVF Conference on Computer Vision and Pattern Recognition (CVPR)},
  pages={6545--6554},
  year={2023}
}

@inproceedings{ni2022expanding,
  title={Expanding language-image pretrained models for general video recognition},
  author={Ni, Bolin and Peng, Houwen and Chen, Minghao and Zhang, Songyang and Meng, Gaofeng and Fu, Jianlong and Xiang, Shiming and Ling, Haibin},
  booktitle={Proceedings of the European Conference on Computer Vision (ECCV)},
  pages={1--18},
  year={2022}
}

@article{snell2017prototypical,
  title={Prototypical networks for few-shot learning},
  author={Snell, Jake and Swersky, Kevin and Zemel, Richard},
  journal={Advances in neural information processing systems},
  volume={30},
  year={2017}
}

@inproceedings{wang2016temporal,
  title={Temporal segment networks: Towards good practices for deep action recognition},
  author={Wang, Limin and Xiong, Yuanjun and Wang, Zhe and Qiao, Yu and Lin, Dahua and Tang, Xiaoou and Van Gool, Luc},
  booktitle={Proceedings of the European Conference on Computer Vision (ECCV)},
  pages={20--36},
  year={2016}
}

@inproceedings{cao2020few,
  title={Few-shot video classification via temporal alignment},
  author={Cao, Kaidi and Ji, Jingwei and Cao, Zhangjie and Chang, Chien-Yi and Niebles, Juan Carlos},
  booktitle={Proceedings of the IEEE/CVF Conference on Computer Vision and Pattern Recognition (CVPR)},
  pages={10618--10627},
  year={2020}
}

@article{zhu2021few,
  title={Few-shot action recognition with prototype-centered attentive learning},
  author={Zhu, Xiatian and Toisoul, Antoine and Perez-Rua, Juan-Manuel and Zhang, Li and Martinez, Brais and Xiang, Tao},
  journal={arXiv preprint arXiv:2101.08085},
  year={2021}
}

@inproceedings{perrett2021temporal,
  title={Temporal-relational crosstransformers for few-shot action recognition},
  author={Perrett, Toby and Masullo, Alessandro and Burghardt, Tilo and Mirmehdi, Majid and Damen, Dima},
  booktitle={Proceedings of the IEEE/CVF Conference on Computer Vision and Pattern Recognition (CVPR)},
  pages={475--484},
  year={2021}
}

@inproceedings{thatipelli2022spatio,
  title={Spatio-temporal relation modeling for few-shot action recognition},
  author={Thatipelli, Anirudh and Narayan, Sanath and Khan, Salman and Anwer, Rao Muhammad and Khan, Fahad Shahbaz and Ghanem, Bernard},
  booktitle={Proceedings of the IEEE/CVF Conference on Computer Vision and Pattern Recognition (CVPR)},
  pages={19958--19967},
  year={2022}
}

@inproceedings{wang2022hybrid,
  title={Hybrid relation guided set matching for few-shot action recognition},
  author={Wang, Xiang and Zhang, Shiwei and Qing, Zhiwu and Tang, Mingqian and Zuo, Zhengrong and Gao, Changxin and Jin, Rong and Sang, Nong},
  booktitle={Proceedings of the IEEE/CVF Conference on Computer Vision and Pattern Recognition (CVPR)},
  pages={19948--19957},
  year={2022}
}

@inproceedings{wu2024efficient,
  title={Efficient few-shot action recognition via multi-level post-reasoning},
  author={Wu, Cong and Wu, Xiao-Jun and Li, Linze and Xu, Tianyang and Feng, Zhenhua and Kittler, Josef},
  booktitle={Proceedings of the European Conference on Computer Vision (ECCV)},
  pages={38--56},
  year={2024}
}

@article{xing2026ma,
  title={MA-FSAR: Multimodal Adaptation of {CLIP} for few-shot action recognition},
  author={Xing, Jiazheng and Zhao, Jian and Xu, Chao and Wang, Mengmeng and Dai, Guang and Liu, Yong and Wang, Jingdong and Li, Xuelong},
  journal={Pattern Recognition},
  volume={169},
  pages={111902},
  year={2026}
}

@inproceedings{hong2021video,
  title={Video pose distillation for few-shot, fine-grained sports action recognition},
  author={Hong, James and Fisher, Matthew and Gharbi, Micha{\"e}l and Fatahalian, Kayvon},
  booktitle={Proceedings of the IEEE/CVF International Conference on Computer Vision (ICCV)},
  pages={9254--9263},
  year={2021}
}

@article{lee2023cast,
  title={CAST: cross-attention in space and time for video action recognition},
  author={Lee, Dongho and Lee, Jongseo and Choi, Jinwoo},
  journal={Advances in Neural Information Processing Systems},
  volume={36},
  pages={79399--79425},
  year={2023}
}

@article{rohrbach15ijcv,
  title={Recognizing fine-grained and composite activities using hand-centric features and script data},
  author={Rohrbach, Marcus and Rohrbach, Anna and Regneri, Michaela and Amin, Sikandar and Andriluka, Mykhaylo and Pinkal, Manfred and Schiele, Bernt},
  journal={International Journal of Computer Vision},
  pages={1--28},
  year={2015}
}

@inproceedings{cao2024task,
  title={Task-Adapter: task-specific adaptation of image models for few-shot action recognition},
  author={Cao, Congqi and Zhang, Yueran and Yu, Yating and Lv, Qinyi and Min, Lingtong and Zhang, Yanning},
  booktitle={Proceedings of the ACM International Conference on Multimedia (ACM MM)},
  pages={9038--9047},
  year={2024}
}

@inproceedings{hmpnet,
author = {Chen, Xiaoyu and Cen, Yigang and Xu, Wanru and Zhang, Yue and Jin, Yi and Li, Yidong and Zhang, Linna},
title = {Hierarchical Meta-prototypes Network for Few-shot Action Recognition},
year = {2025},
isbn = {9798400720352},
publisher = {Association for Computing Machinery},
address = {New York, NY, USA},
url = {https://doi.org/10.1145/3746027.3754703},
doi = {10.1145/3746027.3754703},
booktitle = {Proceedings of the 33rd ACM International Conference on Multimedia},
pages = {2643–2652},
numpages = {10},
location = {Dublin, Ireland},
series = {MM '25}
}

@inproceedings{kumar2025trokens,
  title={Trokens: Semantic-Aware Relational Trajectory Tokens for Few-Shot Action Recognition},
  author={Kumar, Pulkit and Huang, Shuaiyi and Walmer, Matthew and Rambhatla, Sai Saketh and Shrivastava, Abhinav},
  booktitle={2025 IEEE/CVF International Conference on Computer Vision (ICCV)},
  pages={13544--13556},
  year={2025},
  organization={IEEE}
}

@inproceedings{li2025frame,
  title={Frame order matters: A temporal sequence-aware model for few-shot action recognition},
  author={Li, Bozheng and Liu, Mushui and Wang, Gaoang and Yu, Yunlong},
  booktitle={Proceedings of the AAAI Conference on Artificial Intelligence},
  volume={39},
  number={17},
  pages={18218--18226},
  year={2025}
}

@inproceedings{lee2025temporal,
  title={Temporal alignment-free video matching for few-shot action recognition},
  author={Lee, SuBeen and Moon, WonJun and Seong, Hyun Seok and Heo, Jae-Pil},
  booktitle={2025 IEEE/CVF Conference on Computer Vision and Pattern Recognition (CVPR)},
  pages={5412--5421},
  year={2025},
  organization={IEEE}
}

@inproceedings{yang2026mpl,
  title={MPL: Match-guided Prototype Learning for Few-shot Action Recognition},
  author={Yang, Feng and Zhao, Jie and Luo, Fulin and Qin, Anyong and Song, Tiecheng and Zhao, Yue and Gao, Chenqiang and Han, Junwei},
  booktitle={Proceedings of the IEEE/CVF Conference on Computer Vision and Pattern Recognition},
  pages={34439--34448},
  year={2026}
}

@article{cao2026task,
  title={Task-Adapter++: Task-specific adaptation with order-aware alignment for few-shot action recognition},
  author={Cao, Congqi and Han, Peiheng and Zhang, Yueran and Yu, Yating and Lv, Qinyi and Min, Lingtong and Zhang, Yanning},
  journal={Pattern Recognition},
  pages={113237},
  year={2026},
  publisher={Elsevier}
}

@inproceedings{huang2024soap,
  title={SOAP: Enhancing spatio-temporal relation and motion information capturing for few-shot action recognition},
  author={Huang, Wenbo and Zhang, Jinghui and Qian, Xuwei and Wu, Zhen and Wang, Meng and Zhang, Lei},
  booktitle={Proceedings of the 32nd ACM International Conference on Multimedia},
  pages={4572--4580},
  year={2024}
}

@inproceedings{zhu2018compound,
  title={Compound memory networks for few-shot video classification},
  author={Zhu, Linchao and Yang, Yi},
  booktitle={European Conference on Computer Vision (ECCV)},
  pages={782--797},
  year={2018},
  organization={Springer}
}

@inproceedings{fu2020depth,
  title={Depth guided adaptive meta-fusion network for few-shot video recognition},
  author={Fu, Yuqian and Zhang, Li and Wang, Junke and Fu, Yanwei and Jiang, Yu-Gang},
  booktitle={Proceedings of the 28th ACM international conference on multimedia},
  pages={1142--1151},
  year={2020}
}

@inproceedings{wang2021semantic,
  title={Semantic-guided relation propagation network for few-shot action recognition},
  author={Wang, Xiao and Ye, Weirong and Qi, Zhongang and Zhao, Xun and Wang, Guangge and Shan, Ying and Wang, Hanzi},
  booktitle={Proceedings of the 29th ACM international conference on multimedia},
  pages={816--825},
  year={2021}
}

@inproceedings{liu2022task,
  title={Task-adaptive spatial-temporal video sampler for few-shot action recognition},
  author={Liu, Huabin and Lv, Weixian and See, John and Lin, Weiyao},
  booktitle={Proceedings of the 30th ACM International Conference on Multimedia},
  pages={6230--6240},
  year={2022}
}

@inproceedings{wang2023molo,
  title={Molo: Motion-augmented long-short contrastive learning for few-shot action recognition},
  author={Wang, Xiang and Zhang, Shiwei and Qing, Zhiwu and Gao, Changxin and Zhang, Yingya and Zhao, Deli and Sang, Nong},
  booktitle={2023 IEEE/CVF Conference on Computer Vision and Pattern Recognition (CVPR)},
  pages={18011--18021},
  year={2023},
  organization={IEEE}
}

@inproceedings{yu2023multi,
  title={Multi-speed global contextual subspace matching for few-shot action recognition},
  author={Yu, Tianwei and Chen, Peng and Dang, Yuanjie and Huan, Ruohong and Liang, Ronghua},
  booktitle={Proceedings of the 31st ACM International Conference on Multimedia},
  pages={2344--2352},
  year={2023}
}

@inproceedings{zhang2023importance,
  title={On the importance of spatial relations for few-shot action recognition},
  author={Zhang, Yilun and Fu, Yuqian and Ma, Xingjun and Qi, Lizhe and Chen, Jingjing and Wu, Zuxuan and Jiang, Yu-Gang},
  booktitle={Proceedings of the 31st ACM international conference on multimedia},
  pages={2243--2251},
  year={2023}
}

@inproceedings{kumar2024trajectory,
  title={Trajectory-aligned space-time tokens for few-shot action recognition},
  author={Kumar, Pulkit and Padmanabhan, Namitha and Luo, Luke and Rambhatla, Sai Saketh and Shrivastava, Abhinav},
  booktitle={European Conference on Computer Vision (ECCV)},
  pages={474--493},
  year={2024},
  organization={Springer}
}

@inproceedings{herzig2022object,
  title={Object-region video transformers},
  author={Herzig, Roei and Ben-Avraham, Elad and Mangalam, Karttikeya and Bar, Amir and Chechik, Gal and Rohrbach, Anna and Darrell, Trevor and Globerson, Amir},
  booktitle={Proceedings of the IEEE/CVF Conference on Computer Vision and Pattern Recognition (CVPR)},
  pages={3148--3159},
  year={2022}
}

@inproceedings{wang2022bevt,
  title={{BEVT}: Bert pretraining of video transformers},
  author={Wang, Rui and Chen, Dongdong and Wu, Zuxuan and Chen, Yinpeng and Dai, Xiyang and Liu, Mengchen and Jiang, Yu-Gang and Zhou, Luowei and Yuan, Lu},
  booktitle={Proceedings of the IEEE/CVF Conference on Computer Vision and Pattern Recognition (CVPR)},
  pages={14733--14743},
  year={2022}
}

@inproceedings{liu2022video,
  title={Video swin transformer},
  author={Liu, Ze and Ning, Jia and Cao, Yue and Wei, Yixuan and Zhang, Zheng and Lin, Stephen and Hu, Han},
  booktitle={Proceedings of the IEEE/CVF Conference on Computer Vision and Pattern Recognition (CVPR)},
  pages={3202--3211},
  year={2022}
}

@inproceedings{long2025bdc,
  title={{BDC-{CLIP}}: Brownian Distance Covariance for Adapting {CLIP} to Action Recognition},
  author={Long, Fei and Li, Xiaoou and Lv, Jiaming and Yang, Haoyuan and Cheng, Xianjun and Li, Peihua},
  booktitle={Proceedings of the International Conference on Machine Learning (ICML)},
  year={2025}
}

@inproceedings{qing2023disentangling,
  title={Disentangling spatial and temporal learning for efficient image-to-video transfer learning},
  author={Qing, Zhiwu and Zhang, Shiwei and Huang, Ziyuan and Zhang, Yingya and Gao, Changxin and Zhao, Deli and Sang, Nong},
  booktitle={Proceedings of the IEEE/CVF International Conference on Computer Vision (ICCV)},
  pages={13934--13944},
  year={2023}
}

@inproceedings{feichtenhofer2019slowfast,
  title={SlowFast networks for video recognition},
  author={Feichtenhofer, Christoph and Fan, Haoqi and Malik, Jitendra and He, Kaiming},
  booktitle={Proceedings of the IEEE/CVF International Conference on Computer Vision (ICCV)},
  pages={6202--6211},
  year={2019}
}

@inproceedings{zhang2021temporal,
  title={Temporal query networks for fine-grained video understanding},
  author={Zhang, Chuhan and Gupta, Ankush and Zisserman, Andrew},
  booktitle={Proceedings of the IEEE/CVF Conference on Computer Vision and Pattern Recognition (CVPR)},
  pages={4486--4496},
  year={2021}
}

@inproceedings{ao2017quo,
  title={Quo vadis, action recognition? a new model and the kinetics dataset},
  author={Carreira, Joao and Zisserman, Andrew},
  booktitle={Proceedings of the IEEE Conference on Computer Vision and Pattern Recognition (CVPR)},
  pages={21--26},
  year={2017}
}

@inproceedings{zhou2018temporal,
  title={Temporal relational reasoning in videos},
  author={Zhou, Bolei and Andonian, Alex and Oliva, Aude and Torralba, Antonio},
  booktitle={Proceedings of the European Conference on Computer Vision (ECCV)},
  pages={803--818},
  year={2018}
}

@inproceedings{lin2019tsm,
  title={Tsm: Temporal shift module for efficient video understanding},
  author={Lin, Ji and Gan, Chuang and Han, Song},
  booktitle={Proceedings of the IEEE/CVF International Conference on Computer Vision (ICCV)},
  pages={7083--7093},
  year={2019}
}

@article{hendrycks2019benchmarking,
  title={Benchmarking neural network robustness to common corruptions and perturbations},
  author={Hendrycks, Dan and Dietterich, Thomas},
  journal={arXiv preprint arXiv:1903.12261},
  year={2019}
}
}
\clearpage
\newpage
\appendix


\section{Scaling behavior of MATA across capacity, context, and supervision}
\label{sec:scaling}

We analyze how MATA scales with adaptation capacity, temporal context, and supervision, using D$^2$ST as the strongest adapter-based baseline. Across all three axes, MATA consistently exhibits more favorable scaling behavior, indicating that its temporal adaptation mechanism makes more effective use of available resources.

\noindent\textbf{Adaptation depth.}
Table~\ref{tab:blocks_study} shows that MATA outperforms D$^2$ST even when adapters are placed only in the last few transformer blocks, and this margin persists as more blocks are adapted, indicating that MATA does not require deep or widespread architectural modification to model fine-grained temporal structure.

\noindent\textbf{Temporal context.}
From Table~\ref{tab:frames_study}, both methods improve with more input frames, but MATA consistently gains more as clip length grows, suggesting it leverages extended temporal context more effectively to discriminate fine-grained actions.

\begin{table}[h]
\centering
\small
\setlength{\tabcolsep}{5pt}

\begin{subtable}{\linewidth}
\centering
\caption{Comparison of MATA and D$^2$ST when inserting adapters into the last 6, 8, and 10 CLIP blocks. We report 1-shot and 5-shot accuracy on Diving48 and FineGym99.}
\label{tab:blocks_study}
\begin{tabular}{l c c | cc | cc}
\toprule
Method & Blocks & Params &
\multicolumn{2}{c|}{Diving48} &
\multicolumn{2}{c}{FineGym99} \\
& & & 1-shot & 5-shot & 1-shot & 5-shot \\
\midrule
D$^2$ST & 6  & 3.6 & 70.01 & 79.84 & 73.84 & 84.80 \\
MATA    & 6  & \textbf{1.8 }& \textbf{78.43} & \textbf{87.95} & \textbf{79.80} & \textbf{89.60} \\[4pt]
D$^2$ST & 8  & 4.8 & 71.02 & 81.43 & 75.12 & 86.10 \\
MATA    & 8  & \textbf{2.4} & \textbf{79.80} & \textbf{88.02} & \textbf{80.74} & \textbf{90.05} \\[4pt]
D$^2$ST & 10 & 6.1 & 72.70 & 82.64 & 76.32 & 87.13 \\
MATA    & 10 & \textbf{3.0} & \textbf{80.05} & \textbf{88.83} & \textbf{81.66} & \textbf{90.97} \\
\bottomrule
\end{tabular}
\end{subtable}

\vspace{0.5em}

\begin{subtable}{\linewidth}
\centering
\caption{Effect of input frame count on MATA and D$^2$ST. We report 1-shot and 5-shot accuracy using 8, 12, and 16 frames on Diving48 and FineGym99.}
\label{tab:frames_study}
\begin{tabular}{l c | cc | cc}
\toprule
Method & Frames &
\multicolumn{2}{c|}{Diving48} &
\multicolumn{2}{c}{FineGym99} \\
& & 1-shot & 5-shot & 1-shot & 5-shot \\
\midrule
D$^2$ST & 8  & 73.29 & 83.85 & 76.81 & 87.55 \\
MATA    & 8  & \textbf{80.12} & \textbf{90.00} & \textbf{81.37} & \textbf{90.44} \\[4pt]
D$^2$ST & 12 & 79.88 & 87.84 & 80.00 & 89.98 \\
MATA    & 12 & \textbf{84.64} & \textbf{90.68} & \textbf{84.14} & \textbf{92.33} \\[4pt]
D$^2$ST & 16 & 82.75 & 89.82 & 80.41 & 89.38 \\
MATA    & 16 & \textbf{85.90} & \textbf{92.03} & \textbf{88.55} & \textbf{93.76} \\
\bottomrule
\end{tabular}
\end{subtable}
\caption{Ablation on \textbf{adapter depth} and \textbf{temporal context} for MATA and D$^2$ST.}
\label{tab:blocks_frames_combo}
\end{table}

\begin{table*}[ht]

\centering
\small
\setlength{\tabcolsep}{3pt}

\begin{tabular}{l | cccccc | cccccc}
\toprule
Method &
\multicolumn{6}{c|}{Diving48} &
\multicolumn{6}{c}{FineGym99} \\
& 1-shot & 3-shot & 5-shot & 7-shot & 9-shot & 10-shot
& 1-shot & 3-shot & 5-shot & 7-shot & 9-shot & 10-shot \\
\midrule
D$^2$ST &
73.29 & 80.64 & 83.85 & 84.59 & 84.54 & 84.96 &
76.81 & 85.52 & 87.55 & 89.30 & 90.03 & 90.70 \\

MATA &
79.37 & 86.46 & 88.48 & 88.76 & 89.31 & 89.50 &
81.37 & 88.12 & 90.44 & 91.05 & 92.35 & 92.40 \\
\bottomrule
\end{tabular}
\caption{Evaluation of MATA and D$^2$ST across different shot settings. MATA consistently outperforms D$^2$ST in all shot regimes.}
\label{tab:shot_study}
\end{table*}

\noindent\textbf{Supervision.}
Table~\ref{tab:shot_study} shows that MATA maintains a stable advantage over D$^2$ST from 1-shot to 10-shot settings on both datasets, indicating smooth scaling with increased supervision without signs of overfitting.





\section{Nearest-neighbor retrieval across confusing fine-grained classes}
\label{sec:nn-retrieval}

\begin{figure*}[h!]
  \centering
  \includegraphics[width=0.9\linewidth]{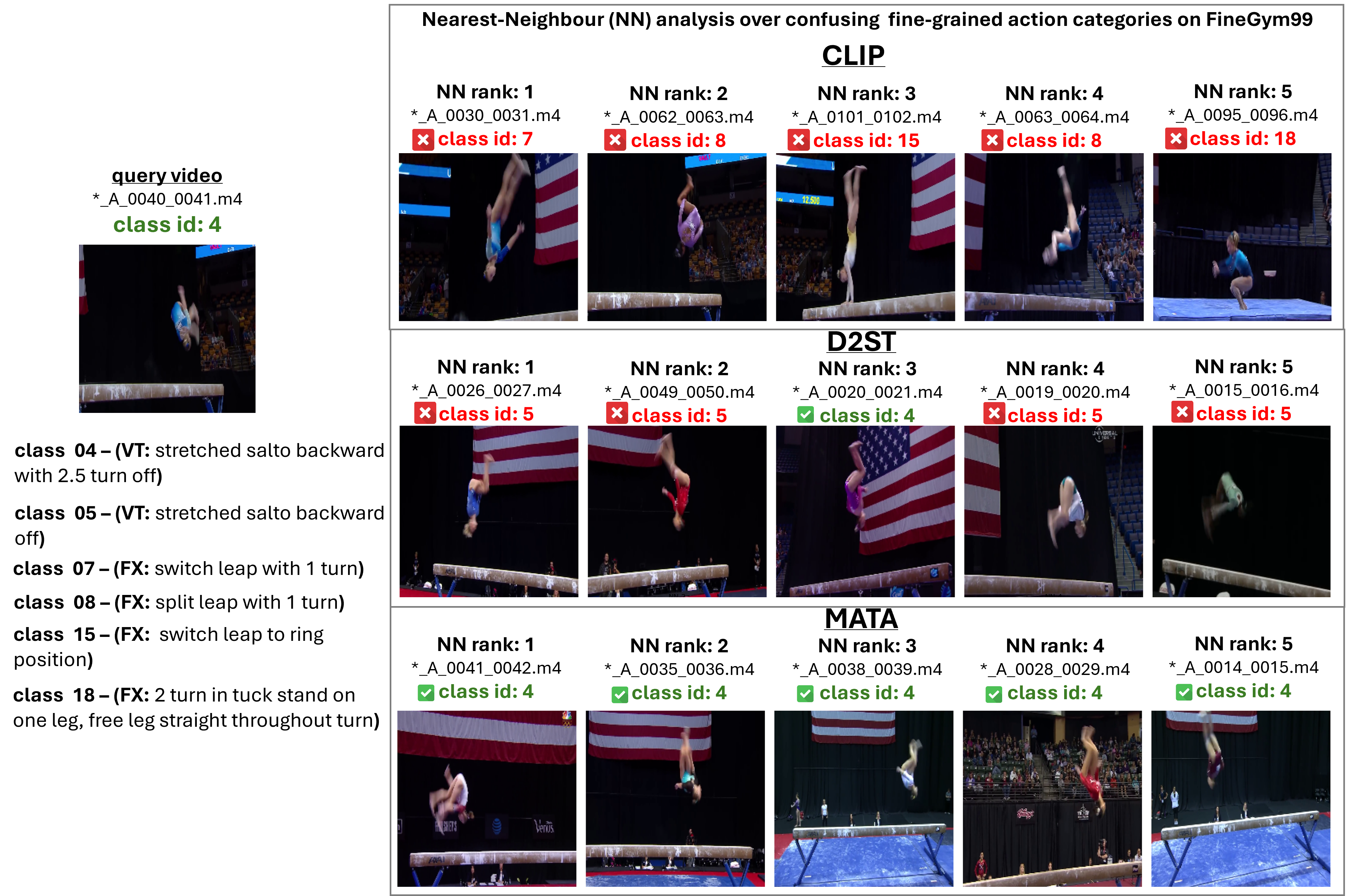}
  \caption{Nearest-neighbor retrieval on confusable FineGym99 classes. For the same query and gallery, MATA yields cleaner top-$K$ neighborhoods than frozen CLIP and D$^2$ST, consistent with improved fine-grained discrimination.}
  \label{fig:NN}
  \vspace{-5mm}
\end{figure*}

We probe how MATA reshapes representation structure using nearest-neighbor (NN) retrieval on confusable FineGym99 classes. We construct a small set of 80 videos from five confusable class pairs and, for each query, retrieve the top-$K$ nearest gallery videos under cosine similarity using three embeddings: (i) frozen CLIP (no adapter), (ii) D$^2$ST, and (iii) MATA. Each video is represented by a single clip embedding $z(v)$ computed with the same 8-frame sampling and preprocessing as episodic evaluation; we $\ell_2$-normalize $z(v)$ so cosine similarity reduces to a dot product. Retrieval results are visualized as grids (Query $\mid$ NN-1..NN-$K$) and summarized in Fig.~\ref{fig:NN}.

To quantify neighborhood quality, we report \emph{Top-5 label purity} (Purity@5), defined as the number of same-label neighbors among the top-5 retrieved videos. Table~\ref{tab:nn-retrieval} shows that MATA produces substantially cleaner neighborhoods than frozen CLIP and D$^2$ST: averaged over 20 queries (5 pairs $\times$ 4 queries per pair), Purity@5 increases from 0.85/5 (frozen CLIP) to 4.00/5 for MATA, with D$^2$ST reaching 2.80/5, confirming MATA's ability to discriminate among fine-grained confusing classes. 
This qualitative and statistical analysis supports our claim that MATA reshapes the local neighborhood structure to better align with discriminative action dynamics rather than appearance alone.

\begin{table}[t]

\centering
\small
\setlength{\tabcolsep}{2pt}

\begin{tabular}{lccc}
\toprule
Method & Purity@5 $\uparrow$ & Purity rate $\uparrow$ & $\Delta$Purity@5 vs. CLIP $\uparrow$ \\
\midrule
Frozen CLIP & 0.85 & 0.17 (0.08--0.27) & 0.00 \\
D$^2$ST        & 2.80 & 0.56 (0.39--0.73) & +1.95 (1.20--2.70) \\
MATA        & 4.00 & 0.80 (0.64--0.93) & +3.15 (2.40--3.90) \\
\bottomrule
\end{tabular}

\caption{\textbf{Nearest-neighbor retrieval quality on FineGym99}. Purity@5 counts how many of the top-5 retrieved neighbors share the query label. Parentheses denote 95\% bootstrap confidence intervals computed over the same 20 queries.}
\label{tab:nn-retrieval}
\end{table}

\section{Single-Frame Blackout Sensitivity Analysis}
\label{sec:blackout}
To understand temporal selectivity, we perform single-frame blackout sensitivity analysis on 8-frame clips from FineGym99.
For each frame $t \in \{1, \ldots, 8\}$, we compute the margin drop $\Delta(t) = m(v) - m(v^{(t)})$, where $m(v) = s_{y^*}(v) - \max_{c \neq y^*} s_c(v)$ is the decision margin for the original clip $v$, $m(v^{(t)})$ is the margin when frame $t$ is replaced with a normalized black frame, $s_{y^*}(v)$ is the cosine similarity score for the ground-truth class $y^*$, and $s_c(v)$ is the score for class $c$.
A large magnitude $|\Delta(t)|$ indicates strong temporal impact, while the sign indicates whether frame $t$ is helpful (positive $\Delta(t)$) or harmful (negative $\Delta(t)$) to the prediction.
We evaluate both MATA and D$^2$ST on 20 randomly sampled 5-way 1-shot episodes from the test split, computing $\Delta(t)$ for each frame and aggregating results across episodes.

\begin{figure*}[t]
    \centering
    \includegraphics[width=0.8\linewidth]{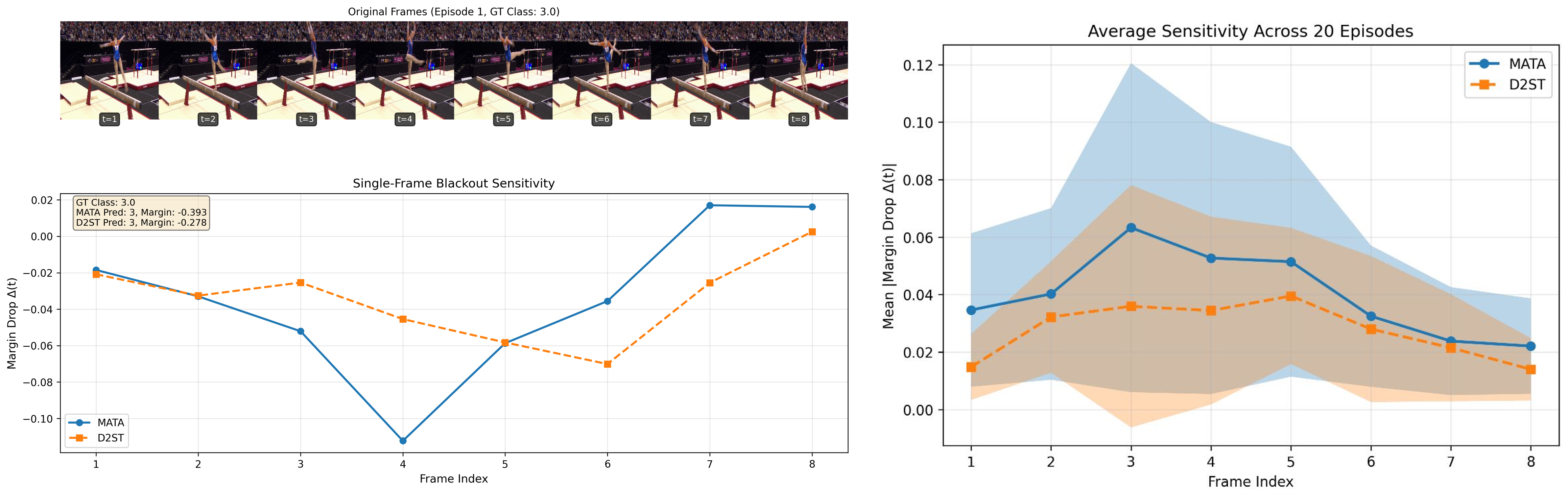}
    \caption{Single-frame blackout sensitivity analysis. (a) Example episode showing MATA's sharp peak at frame 4 ($|\Delta(4)| = 0.112$) vs D$^2$ST's flatter response (peak $|\Delta(6)| = 0.070$) (b) Aggregate plot (mean $|\Delta(t)| \pm$ std over 20 episodes). MATA shows higher peak sensitivity (0.091 vs 0.064) and higher temporal selectivity (0.036 vs 0.025), demonstrating MATA's focused temporal modeling.}
    \label{fig:frame_drop_aggregate}
\end{figure*}

Across 20 episodes, MATA demonstrates higher peak sensitivity (mean $|\Delta(t)|_{\max} = 0.091 \pm 0.054$) and higher temporal selectivity (standard deviation of $\Delta(t)$: $0.036 \pm 0.023$) compared to D$^2$ST ($0.064 \pm 0.038$ and $0.025 \pm 0.014$, respectively).
The aggregate analysis (Figure~\ref{fig:frame_drop_aggregate}) shows MATA consistently identifies key frames (frames 3--4) with sharp peaks, while D$^2$ST exhibits a flatter temporal response across all frames.
A representative example (Figure~\ref{fig:frame_drop_aggregate}, Episode 1) demonstrates MATA's sharp peak at frame 4 ($|\Delta(4)| = 0.112$), indicating strong temporal focus, while D$^2$ST shows a flatter response with lower peak sensitivity ($|\Delta(6)| = 0.070$).
These results demonstrate that MATA's depthwise temporal convolution enables improved temporal selectivity by (1) identifying and focusing on discriminative frames with sharp sensitivity peaks, (2) efficiently distinguishing between critical and non-critical temporal phases, and (3) enabling cheap identification of key frames through margin drop analysis, thereby supporting more efficient temporal modeling compared to attention-based methods that process all frames uniformly.\\

\section{Comparison with other methods on Coarse-grained Datasets}
\label{sec:coarse_results}

Table~\ref{tab:fsar_sota} reports results on coarse-grained benchmarks, where MATA is competitive but does not lead. This is consistent with its design rather than a shortcoming of it. On HMDB51, UCF101, and Kinetics, classes are largely separable from appearance and scene context, and a single well-chosen frame is often sufficient; the discriminative signal these datasets reward is precisely the spatial and semantic content that MATA leaves untouched, since it adapts only the temporal axis. Methods that adapt spatially, or that recruit the CLIP text encoder as an additional supervisory signal such as CLIP-FSAR~\citep{wang2024clip}, MA-FSAR~\citep{xing2026ma}, and Task-Adapter++~\citep{cao2026task}, can exploit that signal directly and hold an advantage here. MATA nevertheless matches or exceeds several of these methods while operating from the video modality alone, without text prompts, pose, depth, or offline preprocessing, and with a fraction of their tunable parameters. Its advantage appears where the discriminative cue is short-lived motion: on the fine-grained benchmarks of Tables~\ref{tab:maincomparison} and~\ref{tab:fewshot_comparison_other_ds}, where classes share actors, scenes, and objects and differ only in the timing and ordering of motion, temporal-only adaptation is the more effective use of a limited parameter budget. We therefore position MATA as a fine-grained specialist rather than a general-purpose few-shot recognizer.

\begin{table*}[ht]
\centering

\resizebox{\textwidth}{!}{%
\begin{tabular}{lllcccccccccc}
\toprule
\multirow{2}{*}{Method} &
\multirow{2}{*}{Reference} &
\multirow{2}{*}{Pretrain} &
\multicolumn{2}{c}{HMDB51} &
\multicolumn{2}{c}{UCF101} &
\multicolumn{2}{c}{Kinetics} &
\multicolumn{2}{c}{SSv2-Small} &
\multicolumn{2}{c}{SSv2-Full} \\
\cmidrule(lr){4-5}
\cmidrule(lr){6-7}
\cmidrule(lr){8-9}
\cmidrule(lr){10-11}
\cmidrule(lr){12-13}
&
&
&
1-shot & 5-shot &
1-shot & 5-shot &
1-shot & 5-shot &
1-shot & 5-shot &
1-shot & 5-shot \\
\midrule

CMN~\cite{zhu2018compound}            & ECCV'18 & ResNet50 & -- & -- & -- & -- & 60.5 & 78.9 & 34.4 & 43.8 & 36.2 & 48.9 \\
OTAM~\cite{cao2020few}          & CVPR'20 & ResNet50 & -- & -- & -- & -- & 73.0 & 85.8 & -- & -- & 42.8 & 52.3 \\
AmeFuNet~\cite{fu2020depth}   & MM'20   & ResNet50 & 60.2 & 75.5 & 85.1 & 95.5 & 74.1 & 86.8 & -- & -- & -- & -- \\
TRX~\cite{perrett2021temporal}        & CVPR'21 & ResNet50 & -- & 75.6 & -- & 96.1 & 63.6 & 85.9 & -- & 59.1 & -- & 64.6 \\
SPRN~\cite{wang2021semantic}         & MM'21   & ResNet50 & 61.6 & 76.2 & 86.5 & 95.8 & 75.2 & 87.1 & -- & -- & -- & -- \\
HyRSM~\cite{wang2022hybrid}       & CVPR'22 & ResNet50 & 60.3 & 76.0 & 83.9 & 94.7 & 73.7 & 86.1 & 40.6 & 56.1 & 54.3 & 69.0 \\
TA$^2$N+Sampler~\cite{liu2022task} & MM'22 & ResNet50 & 59.9 & 73.5 & 83.5 & 96.0 & 73.6 & 86.2 & -- & -- & 47.1 & 61.6 \\
MoLo~\cite{wang2023molo}         & CVPR'23 & ResNet50 & 60.8 & 77.4 & 86.0 & 95.5 & 74.0 & 85.6 & 41.9 & 56.2 & 55.0 & 69.6 \\
MGCSM~\cite{yu2023multi}         & MM'23   & ResNet50 & 61.3 & 79.3 & 86.5 & 97.1 & 74.2 & 88.2 & -- & -- & -- & -- \\
SA-CT~\cite{zhang2023importance}       & MM'23   & ResNet50 & 60.4 & 78.3 & 85.4 & 96.4 & 71.9 & 87.1 & -- & -- & 48.9 & 69.1 \\
CLIP-FSAR~\cite{wang2024clip}& IJCV'24 & ResNet50 & 69.4 & 80.7 & 92.4 & 97.0 & 90.1 & 92.0 & 52.1 & 55.8 & 58.7 & 62.8 \\
SOAP~\cite{huang2024soap}        & MM'24   & ResNet50 & 76.5 & 88.4 & 94.1 & 99.3 & 81.1 & 93.8 & -- & -- & 61.9 & 79.8 \\
HMP-Net      \cite{hmpnet}              & MM'25      & ResNet50 & 77.6 & 88.7 & 94.9 & 99.1 & 92.0 & 94.1 & 54.6 & 60.9 & 61.2 & 76.5 \\
MPL     \cite{yang2026mpl}               & CVPR'26      & ResNet50 & 63.3 & 78.1 & 90.1 & 97.5 & 77.4 & 89.5 & - & - & 43.5 & 56.6 \\
Trokens  \cite{kumar2025trokens}                  & ICCV'25      & DinoV2 & 69.8 & 82.3 & 94.0 & 97.9 & 82.9 & 91.2 & 53.4 & 68.9 & 61.5 & 76.7 \\
TEAM      \cite{lee2025temporal}              & CVPR'25      & ResNet50 & 62.8 & 78.4 & 87.2 & 96.2 & 75.1 & 88.2 & -  & - & - & - \\

\midrule
CLIP-Freeze~\cite{radford2021clip}& ICML'21 & ViT-B & 51.4 & 71.0 & 84.6 & 94.5 & 68.2 & 85.3 & 29.5 & 42.5 & 26.8 & 36.3 \\

CLIP-FSAR~\cite{wang2024clip}* & IJCV'24 & ViT-B & 77.1 & 87.7 & 97.0 & 99.1 & 94.8 & 95.4 & 54.6 & - & - & - \\
MA-FSAR~\cite{xing2026ma}* & PR'26 & ViT-B & 83.4 & 87.9 & 97.2 & 99.2 & 95.7 & 96.0 & 59.1 & 64.5 & 63.3 & 72.3 \\
EMP-Net~\cite{wu2024efficient}*      & ECCV'24 & ViT-B & 76.8 & 85.8 & 94.3 & 98.2 & 89.1 & 93.5 & 57.1 & 65.7 & 63.1 & 73.0 \\
Task-Adapter~\cite{cao2024task}* & MM'24 & ViT-B & 83.6 & 88.8 & 98.0 & 99.4 & 95.0 & 96.8 & 60.2 & 70.2 & 71.3 & 74.2 \\
Task-Adapter++ \cite{cao2026task}*& PR'26 & ViT-B & \textbf{86.5} & \textbf{92.0} & \textbf{98.9} & \textbf{99.7}& \textbf{96.2} &\textbf{ 97.5} & \textbf{63.6} & \textbf{73.2} & \textbf{72.8 }& 75.5 \\

\midrule
SOAP~\cite{huang2024soap}        & MM'24   & ViT-B & 79.3 & 89.8 & 96.8 & 99.5 & 89.9 & 95.5 & -- & -- & 66.7 & 81.2 \\
HMP-Net  ~\cite{hmpnet}                 & MM'25      & ViT-B & 83.8 & 89.9 & 98.7 & 99.5 & 95.7 & 97.5 & 60.5 & 70.4 & 68.4 & 79.5 \\
AIM~\cite{yang2023aim}& ICLR'23 & ViT-B & 74.2 & 86.9 & 95.4 & 98.5 & 88.4 & 95.3 & 52.8 & 67.5 & 63.7 & 79.2 \\
DUALPATH ~\cite{park2023dual}& CVPR'23 & ViT-B & 74.9 & 87.5 & 95.7 & 98.7 & 88.8 & 95.4 & 53.5 & 68.1 & 64.5 & 79.8 \\
ST-Adapter~\cite{pan2022st}& NeurIPS'22 & ViT-B & 74.1 & 87.3 & 95.9 & 98.9 & 88.5 & 95.1 & 53.1 & 68.0 & 64.2 & 79.5 \\
D2ST~\cite{pei2025d2st}& ICCV'25 & ViT-B & 77.1 & 88.2 & 96.4 & 99.1 & 89.3 & 95.5 & 55.0 & 69.3 & 66.7 & \textbf{81.9} \\
TSAM \cite{li2025frame}& AAAI'25 & ViT-B & 84.5 & 88.9 & 98.3 & 99.3 & \textbf{96.2} & 97.1 & 60.5 & 66.7 & 65.8 & 74.6 \\
MPL \cite{yang2026mpl}&  CVPR'26 & ViT-B & 68.0 & 84.9 & 93.0 & 98.5 & 83.7 & 93.8 & - & - & 40.1 & 55.0 \\
TEAM \cite{lee2025temporal}& CVPR'25 & ViT-B & 70.9 & 85.5 & 94.5 & 98.8 & 83.3 & 92.9 & 47.2 & 63.1 & - & - \\
\midrule
MATA                    & --      & ViT-B & 74.10 & 86.54 & 94.82 & 98.50 & - & - & 55.2 & \textbf{73.2} & 58.7 & 74.8 \\

\bottomrule
\end{tabular}}
\caption{\textbf{Comparison with current SOTA few-shot action recognition methods} on HMDB51, UCF101, Kinetics, SSv2-Small, and SSv2-Full datasets. “*” denotes multi-modal methods incorporating additional text input and a text encoder.}
\label{tab:fsar_sota}
\end{table*}

\section{More examples on Temporal Saliency of Adapter Updates}
\label{sec:tempmore}
\begin{figure*}[h!]
  \centering
  \includegraphics[width=0.7\linewidth]{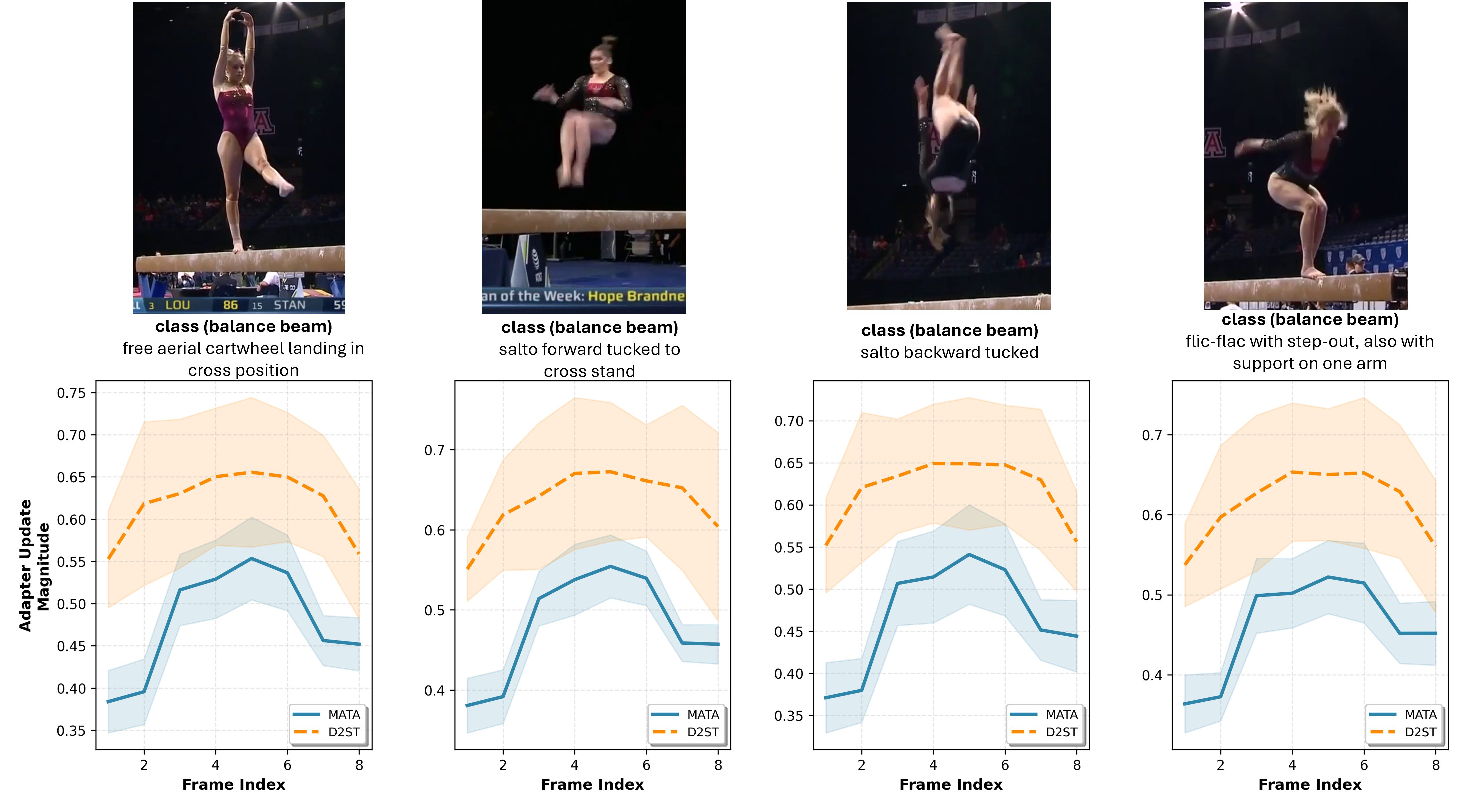}
  \includegraphics[width=.7\linewidth]{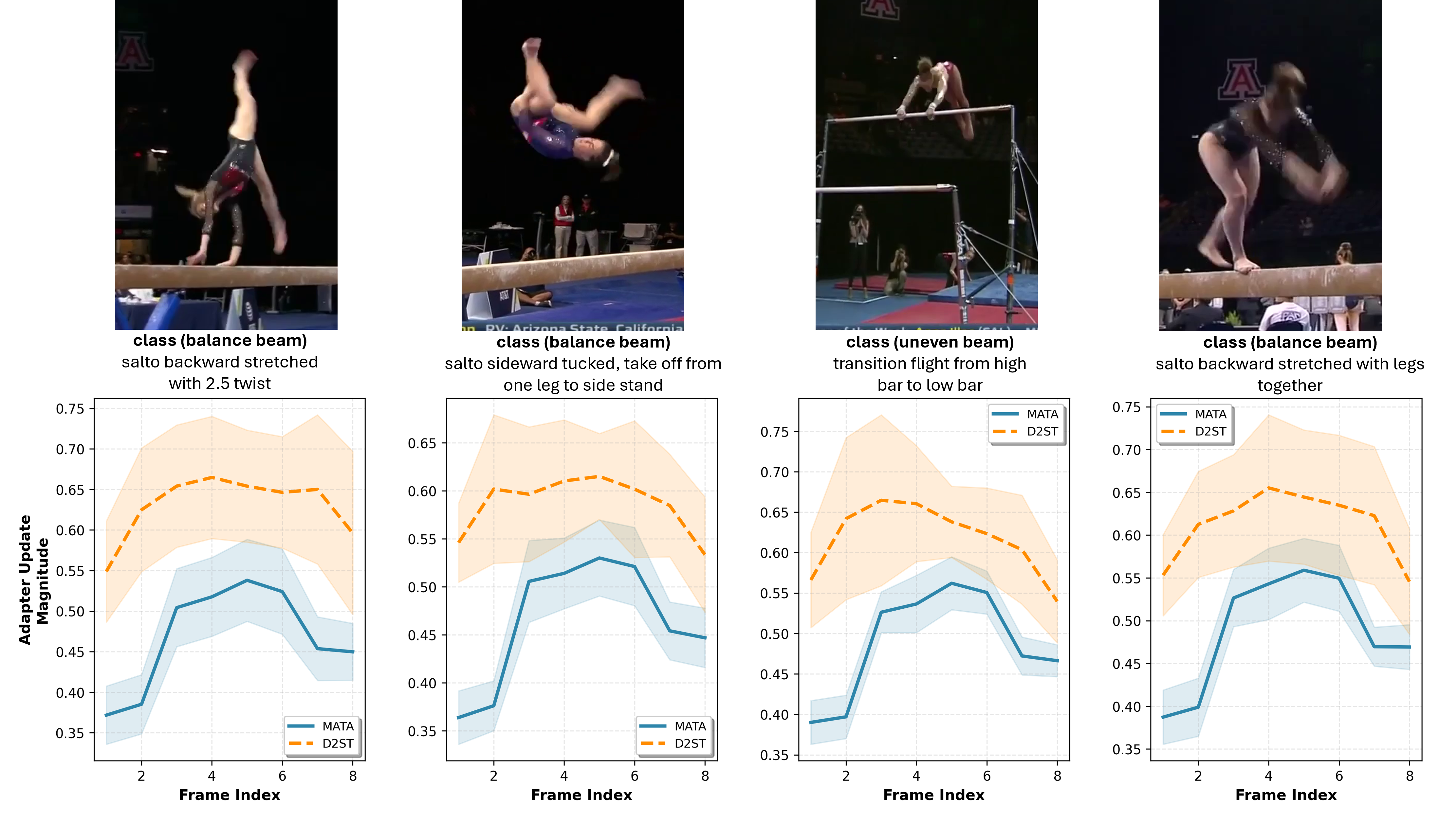}
  \caption{Temporal saliency examples.}
  \label{fig:magnitude2}
\end{figure*}
This section shows more examples for temporal saliency to further strengthen MATA's temporal capacity.

\end{document}